\documentclass[11pt]{article}

\usepackage[preprint]{acl}
\usepackage[table,svgnames,dvipsnames]{xcolor}
\usepackage{fontawesome}
\usepackage{times}
\usepackage{latexsym}
\usepackage[T1]{fontenc}
\usepackage[utf8]{inputenc}
\usepackage{arydshln}
\usepackage{microtype}
\usepackage{pifont}
\usepackage{inconsolata}
\usepackage{graphicx}

\title{\raisebox{-0.45em}{\includegraphics[height=1.6em]{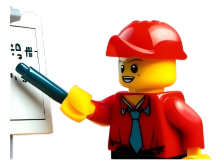}}\, \textsc{AgentAsk}: Multi-Agent Systems Need to Ask}

\author{
  \textbf{Bohan Lin}{\footnotesize $^\bigstar$},
  \textbf{Kuo Yang}{\footnotesize $^\bigstar$}, 
  \textbf{Zelin Tan}{\footnotesize $^{\bigstar,\blacklozenge}$}, 
  \textbf{Yingchuan Lai}{\footnotesize $^\clubsuit$}, 
  \textbf{Chen Zhang}{\footnotesize $^\blacklozenge$},
  \textbf{Guibin Zhang}{\footnotesize $^\spadesuit$}, \\
  \textbf{Xinlei Yu}{\footnotesize $^{\spadesuit}$},
    \textbf{Miao Yu}{\footnotesize $^{\bigstar}$},
  \textbf{Xu Wang}{\footnotesize $^{\bigstar}$}, 
    \textbf{Yudong Zhang}{\footnotesize $^{\bigstar, \text{\faEnvelope}}$}, and
  \textbf{Yang Wang}{\footnotesize $^{\bigstar, \text{\faEnvelope}}$} \\
  {\footnotesize $^\bigstar$}University of Science and Technology of China
  {\footnotesize $^\blacklozenge$}Shanghai AI Laboratory
  \\ 
  {\footnotesize $^\clubsuit$}Xi'an Jiaotong University  
  {\footnotesize $^\spadesuit$}National University of Singapore \\
     {Contact:} \href{mailto:linbohan@stu.xjtu.edu.cn}{\texttt{linbohan@stu.xjtu.edu.cn}},
     \href{mailto:yudong.zhang@ustc.edu.cn}{\texttt{yudong.zhang@ustc.edu.cn}} \\
     {\footnotesize $^{\text{\faEnvelope}}$} {Corresponding authors} \quad \raisebox{-0.4ex}{\includegraphics[height=1em]{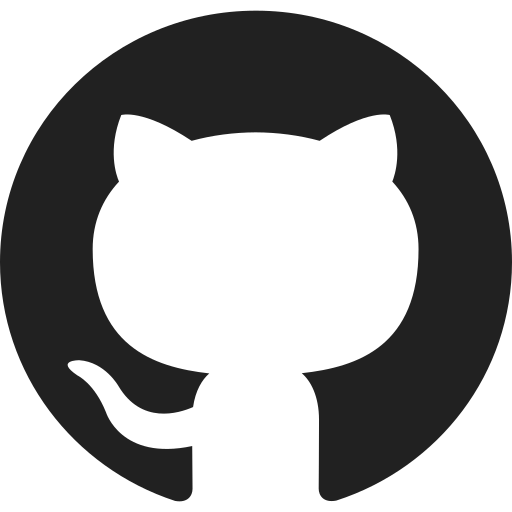}}\hspace{0.3em}\href{https://github.com/BoHan-LIN04/AgentAsk}{\texttt{Code}} 
}

\usepackage{hyperref}
\usepackage{enumitem}
\usepackage{amsmath}
\usepackage{amsfonts}
\usepackage{amssymb}
\usepackage{bbm}
\usepackage{wrapfig} 
\usepackage{url}
\usepackage{colortbl}
\usepackage{mdframed}
\usepackage{float} 
\usepackage{multirow}
\usepackage{subcaption}
\usepackage{booktabs}
\usepackage{algorithm}
\usepackage{algpseudocode}
\usepackage{tabularx}
\usepackage{listings}
\algrenewcommand{\algorithmiccomment}[1]{\hfill\(\triangleright\)\,#1}

\usepackage[nameinlink]{cleveref}
\usepackage{makecell,xcolor,colortbl,tikz}
\newcommand{\inc}[1]{\textcolor{green!60!black}{\(\uparrow\)\,#1}}
\newcommand{\dec}[1]{\textcolor{red!70!black}{\(\downarrow\)\,#1}}

\newcommand{\pmdev}[1]{\textcolor{gray!80!black}{±\,#1}}

\newcommand{\rbdelta}[2]{%
\tikz[baseline=(x.base)]{
  \node[inner sep=0pt] (x) {#1};
  \node[anchor=south east, inner sep=0pt, outer sep=0.1ex,xshift=5ex, yshift=-0.5ex, font=\scriptsize] at (x.south east) {#2};
}
}

\begin{document}
\maketitle
\begin{abstract}
% Multi-agent systems built on large language models (LLMs) promise enhanced problem-solving capabilities through collaborative division of labor. However, they frequently underperform single-agent baselines due to \emph{edge-level} error cascades: minor inaccuracies at one message handoff propagate across the entire chain. We propose \textbf{AgentAsk}, a lightweight and plug-and-play clarification module that treats every inter-agent message as a potential failure point and inserts \emph{minimally necessary} questions to arrest error propagation. AgentAsk follows a three-stage pipeline: (i) distilling edge-level judgments from curated failure traces into a compact policy, (ii) supervising the policy to determine \emph{when/what/whom/how} to ask, and (iii) optimizing online with \textbf{E-GRPO}, a reinforcement learning objective that balances accuracy, latency, and cost. The module is architecture-agnostic and easy to integrate into existing orchestration. Across math, reasoning, and coding benchmarks, AgentAsk consistently improves accuracy and robustness over public multi-agent implementations while keeping overhead minimal, with latency and extra cost all less than 5\%, approaching the performance of a strong evaluator. Beyond empirical improvements, we contribute a principled taxonomy of edge-level errors and a practical recipe for link-local intervention, offering a scalable pathway toward more reliable LLM-based multi-agent systems. 
Multi-agent systems (MAS) built on large language models promise improved problem-solving through collaboration, yet they often fail to consistently outperform strong single-agent baselines due to error propagation at inter-agent message handoffs.
% but often underperform due to error propagation at inter-agent message handoffs.
% However, these systems often underperform compared to single-agent baselines due to error propagation at message handoffs. %We introduce AgentAsk, a lightweight clarification module designed to intervene at the edge level of MAS and prevent error cascades. AgentAsk is grounded in a taxonomy of four edge-level error types: Data Gap, Signal Corruption, Referential Drift, and Capability Gap, which we identify as the sources of failure in multi-agent interactions.
In this work, we conduct a systematic empirical analysis of such failures and introduce an edge-level error taxonomy that identifies four dominant error types: Data Gap, Signal Corruption, Referential Drift, and Capability Gap, as primary sources of failure in multi-agent interactions. 
Building on this taxonomy, we propose AgentAsk, a lightweight clarification module designed to intervene at the edge level in MAS to prevent cascading errors. 
The module operates by strategically applying minimal clarifications at critical points within the system, improving the accuracy and efficiency of the overall task. AgentAsk is trained to balance the trade-offs between clarification cost, latency, and accuracy, while it is also architecture-agnostic and can be easily integrated into existing systems. Evaluated across five benchmarks, AgentAsk consistently improves accuracy by up to 4.69\%, while keeping latency and extra costs below 10\% compared to baseline MAS, showcasing its high efficiency and minimal overhead. 
% This work not only introduces AgentAsk but also contributes a scalable framework for edge-level error correction, offering new insights into improving MAS reliability.
% The code is available at \url{https://anonymous.4open.science/r/AgentAsk-3432}.

\end{abstract}

\section{Introduction}

% Large language model (LLM)–based agent systems have attracted increasing attention for their ability to combine multi-step reasoning with dynamic tool use (\citealp{wei2022chainofthought,schick2023toolformer}). By coordinating multiple LLM-driven agents, these systems aim to leverage collective intelligence to solve complex, real-world tasks that necessitate decomposition, external knowledge access, and iterative refinement\citep{Stanford}. This paradigm is referred to as a multi-agent system (MAS). Applications now span diverse domains such as software engineering (\citealp{metagpt,chatdev}) and AI for Science (\citealp{ghafarollahi2024sciagents,bran2023chemcrow}), and there is growing interest in developing general-purpose multi-agent frameworks for cross-domain problem solving (\citealp{autogen,agentbench,surveyyumiao}). These advances underscore both the promise of multi-agent collaboration and the difficulty of making MAS reliable in practice.
Large language model (LLM)–based agent systems have garnered attention due to their ability to perform multi-step reasoning and dynamic tools. By coordinating multiple LLM-driven agents, these systems leverage collective intelligence to tackle complex tasks that require decomposition, external knowledge access, and iterative refinement \citep{Stanford}. This approach, known as multi-agent systems (MAS), has the potential to solve a wide range of real-world problems \citep{agentbench,surveyyumiao,autogen}.
\begin{figure}
    \centering
    \includegraphics[width=1\linewidth]{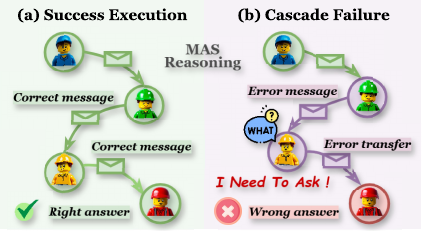}
        \vspace{-0.5cm}
    \caption{\textbf{Multi-Agent System Reasoning: Success vs.\ Cascade Failure.} 
    % The left shows the normal operation of MAS, while the right shows that the upstream agents transmit errors during the interaction that trigger a cascade effect and cause the entire system to fail.
    The left illustrates successful multi-agent reasoning with consistent information exchange, while the right shows how errors introduced by upstream agents propagate across interactions, triggering a cascade that leads to system-level failure.
    }
    \vspace{-0.2cm}
    \label{fig:Taxonomy}
\end{figure}

Despite the promise of collaborative decomposition, multi-agent systems often fail to reliably surpass strong single-agent baselines across diverse tasks. Previous research and benchmarks report unstable performance gains and brittle behavior under realistic orchestration \citep{agentbench,whymasfail}. Execution logs reveal a recurring pattern in which small inconsistencies or missing details introduced at one step travel along the chain and accumulate into system-level failures. Figure~\ref{fig:Taxonomy} illustrates this process. Building on this, chainwise amplification is intensified by specification gaps and inter-agent misalignment, and in some settings, collaboration reduces accuracy \citep{whoandwhen,talk,wang2025comprehensive}. Prior studies quantify the effect, with success on difficult programming tasks falling to about 25\% \citep{whymasfail}. Similar vulnerabilities appear in single-agent traces, including misinterpretations, logical gaps, and limited reflection, which indicates that subtle errors arise early and persist through execution. These observations motivate the problem we study: \textit{how to limit error growth at inter-agent handoffs, such that small local inconsistencies do not accumulate into system-level failures.}

A growing body of recent research attempts to improve the reliability of MAS. One direction emphasizes structured roles and workflow governance, enforcing explicit hierarchies and protocols to reduce miscommunication \citep{metagpt,chatdev,agentverse,yu2025netsafe}. Another line refines orchestration and prompting strategies, leveraging prompt engineering and search to better allocate tasks \citep{gptswarm,aflow,maas,masrouter}. Concurrently, self-checking and feedback mechanisms encourage agents to reflect on or repair their own outputs \citep{debate2,reflection,selfrefine}. These approaches, spanning architecture, algorithm, and introspection, provide useful gains but remain limited. Many proposed solutions are domain-specific, costly to scale, or increase complexity without offering general principles. Moreover, existing taxonomies of failure are largely descriptive and rarely yield prescriptive mechanisms for prevention or correction. Consequently, current systems remain fragile when confronted with complex or open-ended tasks.

% We argue that ensuring reliability necessitates a paradigm shift toward \emph{edge-level} intervention: each inter-agent message should be treated as a potential failure point where minimal clarification can halt cascading errors. To this end, we propose \textbf{AgentAsk}, a plug-and-play clarification module for LLM-based multi-agent systems. AgentAsk introduces a taxonomy of four core edge-level error types and implements a three-stage pipeline: (i) distilling knowledge from failure traces with a powerful evaluator, (ii) transferring this distilled capability to a lightweight clarifier trained to generate targeted questions, and (iii) refining the clarifier through reinforcement learning for adaptive, budget-aware behavior. Crucially, AgentAsk is designed to be lightweight, architecture-agnostic, and easily integrated into existing frameworks, enabling systems to intercept misunderstandings before they spread.
To this end, we argue that ensuring reliability necessitates a paradigm shift toward \emph{edge-level} intervention: each inter-agent message should be treated as a potential failure point where minimal clarification can halt cascading errors. Based on this perspective, we introduce a taxonomy of four core edge-level error types that are responsible for most failures in MAS interactions. Guided by these insights, we propose \textbf{AgentAsk}, a plug-and-play clarification module for LLM-based multi-agent systems. AgentAsk operates by identifying potential failure points at each message handoff and applying minimal clarification to prevent cascading errors.
% achieved through a combination of techniques including knowledge transfer and continuous refinement. 
Crucially, AgentAsk is designed to be lightweight, architecture-agnostic, and easily integrated into existing frameworks, enabling systems to intercept misunderstandings before they spread.

Our work makes the following contributions:
\begin{itemize}[leftmargin=2em,itemsep=-0.1em]
% \item[\ding{182}] \textbf{\textit{Edge-level perspective and taxonomy:}} We formalize a four-type taxonomy of inter-agent errors and motivate clarification as a first-class mechanism for multi-agent collaboration.
\item[\ding{182}] \textbf{\textit{Edge-level perspective and taxonomy:}} 
% We formalize a four-type taxonomy of edge-level inter-agent errors 
We introduce an edge-level perspective on multi-agent failures and formalize a taxonomy of four inter-agent error types,
% and highlight clarification as a key mechanism for improving multi-agent collaboration.
motivating clarification as an effective mechanism for improving multi-agent collaboration.
% \item[\ding{183}] \textbf{\textit{Practical clarification module:}} \textbf{AgentAsk} is a lightweight and plug-and-play module trained with \textbf{E-GRPO} reinforcement learning method, which agents can seamlessly integrate to identify and resolve interaction errors.
\item[\ding{183}] \textbf{\textit{Practical clarification module:}} We propose \textbf{AgentAsk}, a lightweight and plug-and-play clarification module that uses the edge-level taxonomy to identify and resolve interaction errors at message handoffs, improving system reliability with minimal intervention.
\item[\ding{184}] \textbf{\textit{Comprehensive empirical evaluation:}} Extensive experiments across multiple benchmarks show that \textbf{AgentAsk} consistently enhances accuracy by up to \textbf{4.69\%}, while keeping latency and extra cost below 10\% compared to existing multi-agent baselines.
\end{itemize}

\section{Related Work}

\paragraph{LLM-Based Multi-Agent Systems.} MAS coordinates multiple role-specialized agents, often equipped with tools, to solve complex tasks. Early systems demonstrated that organizing multiple LLMs into role-specialized, conversational teams with tool use can outperform single models, through frameworks that scaffold roles, protocols, and conversation programming such as AutoGen~\citep{autogen}, AgentVerse~\citep{agentverse}, CAMEL~\citep{camel}, MetaGPT~\citep{metagpt}, and ChatDev~\cite{chatdev}. Recent work treats orchestration as a learnable object: GPTSwarm models agents and communication links as an optimizable graph \citep{gptswarm}, AFlow searches over code-represented workflows via Monte Carlo tree search \citep{aflow}, MaAS samples query-dependent sub-architectures from an agentic supernet \citep{maas}, and MasRouter learns a cascaded controller for collaboration mode, role allocation, and LLM selection \citep{masrouter}. These directions build on and complement decision-time tool use, such as ReAct~\citep{react} and Toolformer \citep{schick2023toolformer}. Our work is orthogonal to global orchestration search: we hold an existing pipeline fixed and inject a lightweight edge-level clarifier that adds questions at message handoffs to curb error propagation.

% \vspace{-0.2cm}
\paragraph{Failure Analyses in MAS.} Large-scale audits show that multi-agent gains over strong single-agent baselines are not guaranteed and that many failures arise at message handoffs\citep{whymasfail,multiagentbench,whoandwhen}. Methodologically related attempts to improve reliability include multi-agent debate and discussion \citep{debate1,debate2}, self-feedback loops such as Reflexion~\cite{reflection} and Self-Refine \citep{selfrefine}, and verification or self-correction procedures \citep{cannotselfcorrect,revise}, yet recent analyses document conditions where debate or naive self-correction can fail or even degrade accuracy \citep{talk}. A complementary line advocates asking clarifying questions under uncertainty in single-agent human–computer interaction\citep{awn,futureask,rlclarify,askclar}. We follow this clarification-centric view but operationalize it for multi-agent settings with an edge-aware policy that decides when, what, whom, and how to ask exactly at the handoff.

\section{Error Taxonomy}
\vspace{-0.10cm}
\label{taxonomy}
\textbf{Conceptual Inspiration.}
In multi-agent systems, ensuring reliable communication between agents is essential to prevent cascading errors that can propagate throughout the system. Our error taxonomy is inspired by human collaboration theory \citep{GittellJH2006RcCw,human1}, including the concept of \emph{Relational Coordination}, which emphasizes the importance of frequent, timely, and accurate communication, shared goals, and mutual respect among team members. According to this theory, coordination can falter when communication is unclear, roles are misaligned, or team members fail to anticipate each other's needs, leading to errors in the system.

\begin{figure}
    \centering
    \includegraphics[width=1\linewidth]{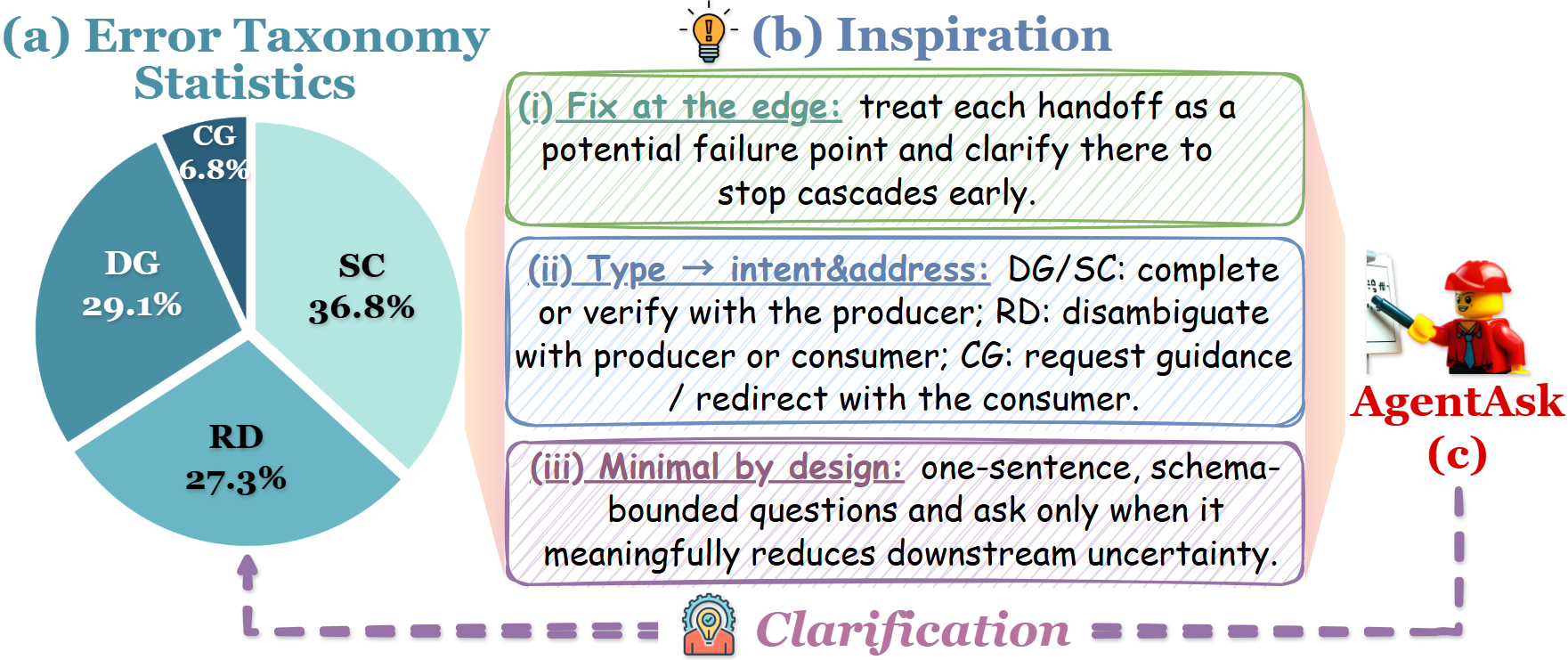}
    \caption{\textbf{From taxonomy to design.} (a) Empirical distribution of edge-level error types ($N$=824). (b) Three design inspirations distilled from our taxonomy: localize fixes at the edge; map \emph{type}→\emph{intent \& addressee}; enforce minimality-by-design. (c) Our designed AgentAsk for edge-level intervention (see Section~\ref{sec:Methodology}).}
    \vspace{-0.5cm}
    \label{fig:pie}
\end{figure}

\noindent \textbf{Empirical Evidence.} In this work, we empirically demonstrate that chain-style error propagation is a fundamental root cause of failures in MAS, where a single error can cascade into a system-wide collapse. We reveal these errors through an audit of \textbf{\underline{824}} execution logs, annotating each agent-to-agent message as the unit of analysis. To ensure the reliability of our classification, the annotation process was conducted by multiple professional annotators with expertise in MAS. These annotators engaged in several rounds of cross-annotation and adjudication to achieve consistent results. The high inter-rater reliability was confirmed with a Fleiss' Kappa value of \textbf{\underline{0.84}}, indicating substantial agreement among annotators. Our empirical results, shown in Figure~\ref{fig:pie}, indicate that most errors in MAS stem from violations of a fundamental principle: \textit{agents should complete tasks \textbf{accurately} within \textbf{their capabilities} and deliver \textbf{clear and complete information} handovers to downstream agents \textbf{without omission}.} Building on this observation, four types of errors have been condensed.

% By focusing on these edge-level communication points, we propose a taxonomy that allows for targeted interventions to prevent small errors from escalating into larger system failures. The ability to identify and address these errors at the point of handoff ensures that each agent's capabilities are respected, and that information flows clearly and completely throughout the system.

\subsection{Data Gap}
In a good collaboration, a handoff never misses a crucial detail. In human teams, a brief that forgets a constraint sends downstream work off course, and the cost multiplies with each relay. In multi-agent settings, the same dynamic appears when a boundary case, a unit, or a condition is left out. Small omissions turn into systemic bias as later steps harden a wrong default. A short check at the handoff that lists required fields and confirms ranges stops the slide before it spreads.

\subsection{Referential Drift}
Teams falter when words stop pointing to the same thing. People know the feeling when two teammates use different names for the same object, and the dialogue quietly splits into parallel threads. Agents replicate this when pronouns, variable names, or record keys are passed without grounding, so later steps bind them inconsistently. The result is confident but incompatible reasoning. A crisp confirmation of who is who and which value is which at the handoff restores a single shared state.

\subsection{Signal Corruption}
A distorted intermediate result enters the chain and gets propagated as truth. In human work, a spreadsheet with a subtle formula error can contaminate every chart that follows. Agents exhibit the same fragility when a wrong fact, a malformed structure, or a unit mistake is presented with fluency and then reused without scrutiny. Downstream modules amplify the error instead of correcting it. A targeted request to verify the source value or structure at the handoff contains the damage early.

\subsection{Capability Gap}
    \vspace{-0.15cm}
Healthy organizations respect the limits of each role. Teams recognize this when a specialist is asked to deliver outside their training, and improvisation replaces method. In multi-agent pipelines, this appears when a role tuned for explanation is asked for algebra or when a planner is pushed into domain judgment. Progress stalls because the wrong skill sits at the critical edge. An early redirect or a brief request for guidance at the handoff aligns the task with a capable peer and keeps the chain coherent.

\begin{figure*}[t]
    \centering
    \includegraphics[width=1\linewidth]{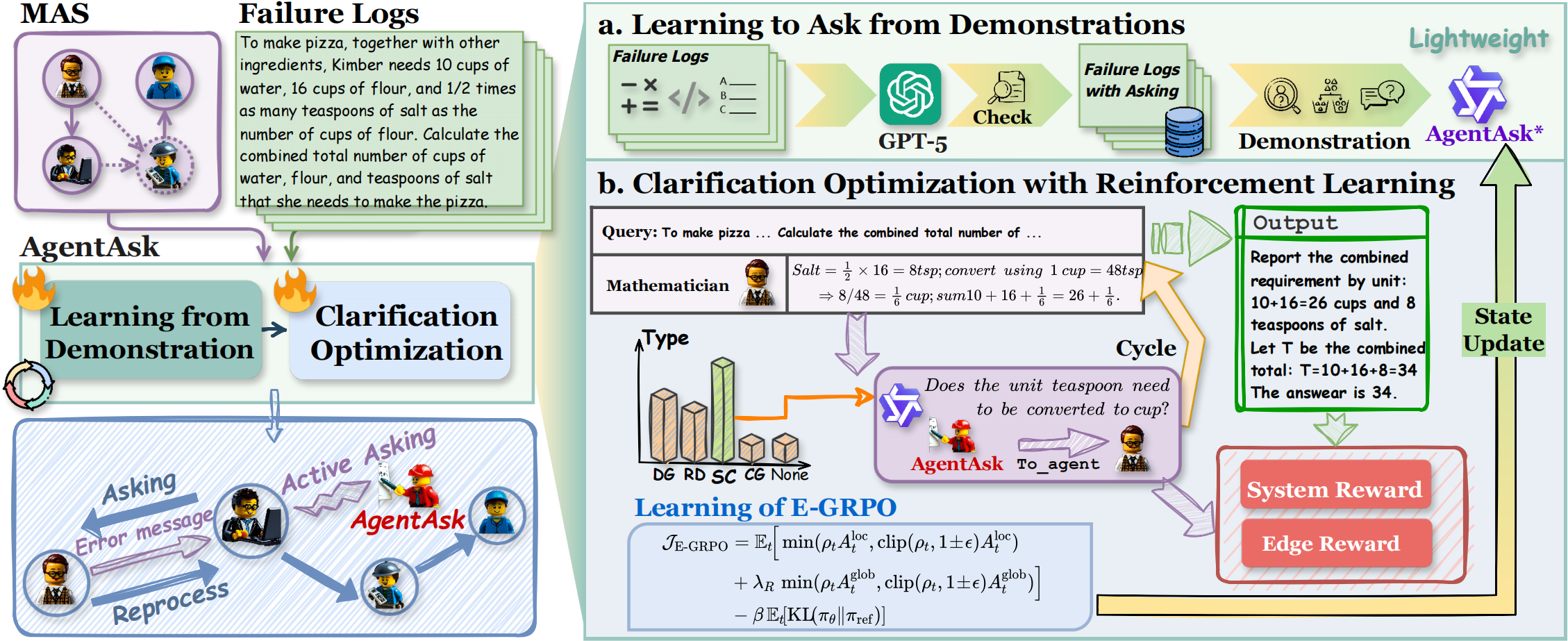}
            \vspace{-0.4cm}
    \caption{\textbf{Overview of AgentAsk.} The module operates at the \emph{edge level}, treating each inter-agent message as a potential failure point. The figure illustrates both the architecture and training process: (i) Constructing an edge-level corpus from failure logs by a large evaluator, (ii) Learning to ask from demonstrations to decides \emph{when/what/whom/how} to ask, and (iii) reinforcement learning with E\textit{-}GRPO for clarification optimization under latency and cost constraints. This pipeline equips AgentAsk with edge-aware monitoring and minimal, targeted interventions while remaining architecture-agnostic and easy to integrate into diverse MASs.}
        \vspace{-0.3cm}
    \label{fig:framework}
\end{figure*}

    \vspace{-0.1cm}
\subsection{Inspiration from Edge-level Taxonomy}
    \vspace{-0.15cm}
Our analysis shows that all four error types manifest as failures at the edge, the information handoff between agents in a multi-agent system. This shifts the design focus to the handoff itself rather than the global workflow. \textbf{In our annotated corpus ($N$=824), these errors account for most failures,} with Data Gap 29.1\%, Referential Drift 27.3\%, Signal Corruption 36.8\%, and Capability Gap 6.8\%. According to the human collaboration theory we mentioned above, effective handoffs are governed by three habits: \ding{182} check the handoff for missing content, ambiguous reference, corrupted intermediate results, or role mismatch; \ding{183} if a problem exists, fix it at the source so it does not propagate downstream; \ding{184} intervene only as needed, keeping the exchange brief and within latency and cost limits. 

These habits motivate an edge-centered design that monitors each transfer and issues minimal clarifications exactly at the point of exchange. The illustrative examples can be seen in Appendix~\ref{app:case}.

    \vspace{-0.2cm}
\section{Design of AgentAsk}
\label{sec:Methodology}
    \vspace{-0.2cm}
\textbf{Overview.} Grounded in the taxonomy, failures cluster at the edge where one agent hands information to the next. We therefore design around three moves and introduce \textbf{AgentAsk}, a framework for real-time monitoring and correction in multi-agent systems. First, to identify errors, we place an external clarifier at each handoff that inspects the outgoing message and its local context, flagging risks early. Second, to correct errors, we use “Ask”: the AgentAsk sends a brief, targeted question to the right party so the handoff is repaired before the chain advances. Third, we transferred the asking capabilities of the powerful model to AgentAsk, but to keep correction efficient, fine-tuning alone is not enough, so we add reinforcement learning that adapts the ask decision, the addressee, and the phrasing to context and budget. 
% Concretely, we use a two-stage training paradigm, \emph{Knowledge Transfer via Supervised Fine-tuning} followed by \emph{Autonomous Learning via Reinforcement Optimization}, which gives \textbf{AgentAsk} an inherited ability to detect faulty outputs and a learned ability to apply minimal, cost-aware clarifications. 
% The details of the framework training are explained in detail in the following chapters. 
% The empirical experiments are provided in Section~\ref{sec:experiments}. 

The training process of the framework is detailed in the following chapters, while its overall design is outlined in Figure~\ref{fig:framework}. Complete training pseudocode appears in Appendix~\ref{app:A} as Algorithm~\ref{alg:agentask}, and the notation used throughout is summarized in Table~\ref{tab:notations} of Appendix~\ref{sec:notations}.

\subsection{Clarification as Edge Intervention}
\label{sec:modeling}

We model multi-agent collaboration as a sequence of message handoffs over a directed interaction graph. At each step, the controller \textbf{AgentAsk} receives a local view of the exchange and decides whether to insert a clarification to improve task outcomes.
Each decision is based on an edge-local state $x_t$, where $t$ is the timestep of the handoff, and results in an action $a_t = (z_t, \tilde{v}_t, q_t)$, where $z_t$ is a binary gate indicating whether to ask, $\tilde{v}_t$ selects which agent to query, and $q_t$ is the generated clarification question. The controller either allows the message to proceed or poses a brief clarification to one of the agents. This intervention is lightweight but can substantially improve task-level accuracy.

The policy $\pi_\theta$ governs the action sequence across an input, yielding a trajectory $\tau = \{(x_t, a_t)\}_{t=1}^T$. The controller is optimized to maximize expected task utility while operating under a cost constraint:
\begin{equation}
\label{eq:constrained_obj}
\max_{\theta}\; \mathbb{E}_{\tau \sim \pi_\theta} \left[ U(\tau) \right] \quad \text{s.t.} \quad \mathbb{E} \left[ C(\tau) \right] \le B.
\end{equation}
Here, $U$ is the utility function measuring task success, and $C$ represents the cost incurred from clarifications, while $B$ is the predefined budget limit.
The internal structure of $x_t$ and $a_t$, as well as the factorized form of $\pi_\theta$, are introduced and discussed in Appendix~\ref{app:controller_details}.

\subsection{Learning to Ask from Demonstrations}

We initialize the controller under a cold-start condition using fine-tuning on curated clarification traces. This allows AgentAsk to acquire structured ask behavior before reinforcement optimization.
We construct a dataset of edge-local examples, each consisting of a context $x_i$ and an intervention $y_i$. Specifically, $x_i$ represents the state at edge $i$, and $y_i$ encodes the teacher's decision (whether to ask, whom to ask, and what to ask). The controller learns to predict the same content.

Let $\mathcal{D} = \{(x_i, y_i)\}_{i=1}^{N}$ be the training dataset, where $N$ is the total number of examples. The training loss includes a decision component $\mathcal{L}_{\text{type}}$ for predicting whether a clarification is needed, and a conditional ask component $\mathcal{L}_{\text{ask}}$ for predicting the clarification details:
\begin{equation}
\label{eq:sft_total}
\mathcal{L}_{\text{Total}} = \mathcal{L}_{\text{type}} + \lambda_{\text{ask}}\, \mathcal{L}_{\text{ask}}.
\end{equation}
Here, $\lambda_{\text{ask}}$ is a parameter that balances two components.
This provides a cold-start policy capable of clarification. Details of the data construction, label format, and loss terms appear in Appendix~\ref{app:sft_loss}.

\subsection{Clarification Optimization with Reinforcement Learning}
% Supervised fine-tuning transfers edge-level judgment and question quality from a strong evaluator, but it only imitates the teacher on logged cases. Imitation is static: it is limited by dataset coverage, does not learn when not to ask, and cannot balance accuracy against constraints under a new scenario. The ask gate and addressee are discrete choices whose value is revealed only after downstream effects, so credit must flow across the chain. We therefore continue with reinforcement learning to adapt the factored controller in Eq.~(2) to the constrained objective in Eq.~(3), using shaped rewards that turn these trade-offs into learnable signals.
Previous training enables us to transfer edge-level judgment and question quality from a strong evaluator. However, imitation learning has its limits: it only captures behavior from logged cases and cannot learn when not to ask, nor can it balance accuracy against constraints in new scenarios. The ask gate and addressee are discrete decisions whose true value emerges only after downstream effects, so credit must propagate through the entire sequence. Therefore, we continue with reinforcement learning to adapt the $\pi_{\theta}$ to the constrained objective in Equation~\eqref{eq:constrained_obj}, using shaped rewards that convert these trade-offs into learnable signals.
\vspace{-0.2cm}
\paragraph{Reward Design.} First, we encourage asking only when it actually removes downstream uncertainty:
\begin{equation}
\label{eq:rl_eff}
r_t^{\mathrm{eff}}=
\begin{cases}
\ \,1, & z_t=1 \wedge n_{t+1}=0,\\
-1, & z_t=1 \wedge n_{t+1}=1,\\
\ \,0, & z_t=0.
\end{cases}
\end{equation}
Here, $n_{t+1}$ represents whether need to clarify again, that AgentAsk may ask multiple times at one edge.

Second, we enforce parsimony to keep clarifications scarce via a sliding counter $c_t$:
\begin{equation}
\label{eq:rl_par}
r_t^{\mathrm{par}}=-\lambda_{\mathrm{sw}}\max(c_t-1,0).
\end{equation}
To stabilize outputs and preserve brevity, we introduce a format bonus:
\begin{equation}
\label{eq:rl_fmt}
r_t^{\mathrm{fmt}}=\alpha_{\mathrm{fmt}}\mathbbm{1}[F_t=1].
\end{equation}
These signals are aggregated into an edge reward that embodies our inspirations from taxonomy,
\begin{equation}
\label{eq:rl_edge}
r_t^{\mathrm{edge}}=\alpha_{\mathrm{eff}}\,r_t^{\mathrm{eff}}+r_t^{\mathrm{par}}+r_t^{\mathrm{fmt}},
\end{equation}
while a terminal score anchors final correctness,
\begin{equation}
\label{eq:rl_sys}
R=\alpha_{\mathrm{ans}}\,s.
\end{equation}
% Together, Eqs.~\eqref{eq:rl_edge}–\eqref{eq:rl_sys} align the controller with our objective: intervene only when a concise question can halt DG/RD/SC/CG propagation, avoid gratuitous turns, keep outputs well-formed, and ultimately solve the task.
where $s$ is the final success score of the task.
\vspace{-0.2cm}
\paragraph{Optimization: E-GRPO.}
% Before the episode reaches its terminal agent, the final correctness reward $R$ is unavailable; By following the Group Relative Policy Optimization (GRPO)~\citep{grpo}, in this prefix regime, we train \emph{exactly} on the edge-level signal by applying a within-edge relative update that selects the best clarification using $r_t^{\mathrm{edge}}$. Once the episode terminates and $R$ is observed, we \emph{augment} the same update with a global credit that allocates outcome utility back to each edge occurrence, thereby coupling local containment with end-task success.
We extend the principles of Group Relative Policy Optimization (GRPO)~\citep{grpo} by introducing a \textit{gradual global credit mechanism} to be our \textbf{E\textit{-}GRPO}. While GRPO optimizes policies based on edge-level rewards, we ensure that \textbf{E\textit{-}GRPO aligns local decisions with overall success by incorporating global feedback in a controlled manner.}

For each edge $t$, we form a local advantage $A_{t}^{\mathrm{loc}} = r_t^{\mathrm{edge}}$ by the controller. The system-level signal $A_t^{\mathrm{glob}}=w_t\,(R-b)$ is distributed at time $t$ with nonnegative weights $w_t$ and an action-independent baseline $b$. The weight $w_t$ increases as the training progresses, ensuring that global feedback starts to influence the policy more as the agent gains confidence. Initially, the agent learns from edge-level rewards $r_t^{\mathrm{edge}}$, which guide the local decisions. However, to prevent the agent from getting trapped in local optima, we introduce global credit to reflect the overall success of the task. This global credit is gradually integrated into the learning process through a time-varying weight $\lambda_R(t)$, which increases as the training progresses.

Let $\rho_{t}=\pi_\theta(a_{t}\mid h_{t})/\pi_{\mathrm{old}}(a_{t}\mid h_{t})$. We maximize the E-GRPO surrogate that shares ratios for local and global credit and regularizes toward a reference:
\begin{equation}
\label{eq:rl_obj}
\resizebox{0.95\columnwidth}{!}{$
\begin{aligned}
\mathcal{J}_{\mathrm{E\text{-}GRPO}}(\theta)
= &\mathbb{E}_{t}\Big[
      \min\!\big(\rho_{t}A_{t}^{\mathrm{loc}},\,
                 \mathrm{clip}(\rho_{t},1\!\pm\!\epsilon)\,A_{t}^{\mathrm{loc}}\big)
\\
&\quad
    +\lambda_R\,\min\!\big(\rho_{t}A_t^{\mathrm{glob}},\,
                 \mathrm{clip}(\rho_{t},1\!\pm\!\epsilon)\,A_t^{\mathrm{glob}}\big)
\Big]
\\
&
 -\,\beta\,\mathbb{E}_{t}\!\left[
     \mathrm{KL}\big(\pi_\theta(\cdot\mid h_t)\,\|\,\pi_{\mathrm{ref}}(\cdot\mid h_t)\big)
 \right].
\end{aligned}
$}
\end{equation}
When $R$ is not available, the second term in Eq.~\eqref{eq:rl_obj} is simply absent, and training proceeds purely on edge-level selection; when $R$ arrives, it is injected uniformly through $A_t^{\mathrm{glob}}$, ensuring that \textit{minimal, successful clarifications remain aligned with end-to-end correctness} while updates stay conservative through clipping and reference KL regularization.
% (where “KL” is the Kullback–Leibler divergence between the current edge policy and the initialized reference, acting as a trust-region regularizer that preserves the clarifier’s brevity and schema prior).

% This approach ensures that \textbf{AgentAsk} learns to preventing it from becoming stuck in local optima while progressively aligning local actions with global outcomes.
This approach ensures that \textbf{AgentAsk} avoids becoming trapped in local optima, while progressively aligning edge-level decisions with global task outcomes,
thereby improving the reliability of multi-agent systems by preventing early error propagation with minimal overhead.

\begin{table*}[!t]
\centering
\small
\renewcommand{\arraystretch}{1.2}
\setlength{\tabcolsep}{6pt}
\begin{tabular}{llcccccc}
\Xhline{1.2pt}
\rowcolor{CadetBlue!16}
\textbf{Method} &\textbf{Settings} & \textbf{GSM8K} & \textbf{MATH} & \textbf{HumanEval} & \textbf{MMLU} & \textbf{MBPP} & \textbf{Average} \\
\hline\hline
\rowcolor{gray!10}
IO & & 89.52 & 49.23 & 89.66 & 79.63 & 71.38 & 75.88 \\
CoT & & 89.71 & 48.19 & 90.64 & 80.44 & 70.66 & 75.93 \\
\rowcolor{gray!10}
Self-Refine & & 90.37 & 47.92 & 88.88 & 78.31 & 71.49 & 75.40 \\
\hline
\multirow{3}{*}{GPTSwarm}
  & origin & 92.18 & 51.41 & 90.63 & 81.26 & 76.81 & 78.46 \\
  & \texttt{+GPT-5} & \rbdelta{95.55}{\pmdev{1.42}} & \rbdelta{56.30}{\pmdev{1.15}} & \rbdelta{94.05}{\pmdev{0.88}} & \rbdelta{86.85}{\pmdev{1.73}} & \rbdelta{82.45}{\pmdev{1.56}} & 83.04\inc{(+4.58)} \\
  & \textbf{+\textsl{AgentAsk}} & \rbdelta{\textbf{94.52}}{\pmdev{1.27}} & \rbdelta{\textbf{55.05}}{\pmdev{1.64}} & \rbdelta{\textbf{92.95}}{\pmdev{1.09}} & \rbdelta{\textbf{85.25}}{\pmdev{1.81}} & \rbdelta{\textbf{81.15}}{\pmdev{1.33}} & \textbf{81.78}\inc{(\textbf{+3.32})} \\
\hline
\rowcolor{gray!10}
  & origin & 91.35 & 50.89 & 91.67 & 82.16 & 79.63 & 79.14 \\
  \rowcolor{gray!10}
  \multirow{-1}{*}{AFlow}
  & \texttt{+GPT-5} & \rbdelta{94.85}{\pmdev{1.68}} & \rbdelta{56.20}{\pmdev{0.92}} & \rbdelta{95.05}{\pmdev{1.47}} & \rbdelta{87.75}{\pmdev{1.25}} & \rbdelta{85.10}{\pmdev{1.80}} & 83.79\inc{(+4.65)} \\
  \rowcolor{gray!10}
  & +\textsl{\textbf{AgentAsk}} & \rbdelta{\textbf{93.70}}{\pmdev{1.54}} & \rbdelta{\textbf{54.95}}{\pmdev{1.38}} & \rbdelta{\textbf{93.95}}{\pmdev{1.12}} & \rbdelta{\textbf{86.40}}{\pmdev{1.66}} & \rbdelta{\textbf{83.90}}{\pmdev{1.05}} & \textbf{82.58}\inc{(\textbf{+3.44})} \\
\hline
\multirow{3}{*}{MaAS}
  & origin & 92.84 & 51.08 & 92.02 & 82.93 & 81.22 & 80.02 \\
  & \texttt{+GPT-5} & \rbdelta{96.20}{\pmdev{1.89}} & \rbdelta{56.85}{\pmdev{1.51}} & \rbdelta{95.45}{\pmdev{0.76}} & \rbdelta{88.35}{\pmdev{1.94}} & \rbdelta{86.70}{\pmdev{1.37}} & 84.71\inc{(+4.69)} \\
  & +\textsl{\textbf{AgentAsk}} & \rbdelta{\textbf{95.10}}{\pmdev{1.23}} & \rbdelta{\textbf{55.55}}{\pmdev{1.70}} & \rbdelta{\textbf{94.45}}{\pmdev{1.58}} & \rbdelta{\textbf{86.95}}{\pmdev{1.44}} & \rbdelta{\textbf{85.30}}{\pmdev{1.19}} & \textbf{83.47}\inc{(\textbf{+3.45})} \\
\hline
\rowcolor{gray!10}
  & origin & 93.26 & 51.52 & 91.03 & 83.19 & 79.04 & 79.61 \\
  \rowcolor{gray!10}
  \multirow{-1}{*}{MasRouter}
  & \texttt{+GPT-5} & \rbdelta{95.10}{\pmdev{1.05}} & \rbdelta{56.70}{\pmdev{1.82}} & \rbdelta{94.55}{\pmdev{1.31}} & \rbdelta{88.55}{\pmdev{1.67}} & \rbdelta{85.15}{\pmdev{1.96}} & 84.01\inc{(+4.40)} \\
  \rowcolor{gray!10}
      & +\textsl{\textbf{AgentAsk}} & \rbdelta{\textbf{94.86}}{\pmdev{1.48}} & \rbdelta{\textbf{55.65}}{\pmdev{1.29}} & \rbdelta{\textbf{93.35}}{\pmdev{1.74}} & \rbdelta{\textbf{87.10}}{\pmdev{1.52}} & \rbdelta{\textbf{83.55}}{\pmdev{1.08}} & \textbf{82.54}\inc{(\textbf{+3.29})} \\
\Xhline{1.2pt}
\end{tabular}
\caption{\textbf{Accuracy/Pass@1 across frameworks under three settings (\texttt{origin}, \texttt{+GPT-5}, \texttt{+AgentAsk}) on five benchmarks.} For improved settings (+GPT-5 and +AgentAsk), bottom-right overlays show standard deviation in gray. Average column shows overall percentage-point gains (green \textcolor{green!60!black}{$\uparrow$}) relative to each framework’s \texttt{origin}.}
\vspace{-0.3cm}
\label{tab:main}
\end{table*}

\vspace{-0.20cm}
\section{Experiments}
\label{sec:experiments}
\vspace{-0.20cm}
% We evaluate whether a lightweight \emph{edge-level} clarifier can enhance the reliability of multi-agent systems by intervening at agent-to-agent handoff points, without altering the original orchestration. Our experiments validate the edge-level clarification paradigm, quantify its overhead in latency and cost, and analyze its effectiveness through a taxonomy of error types. %Additionally, we explore whether the clarifier can progressively improve its error-catching ability and become more efficient through semi-off-policy online learning. 
% To ensure broad coverage and fair comparison, we test our method across diverse benchmarks in math reasoning, general-knowledge QA, and code generation, integrating it into multiple representative multi-agent frameworks.  We compare our lightweight \texttt{AgentAsk} with a strong clarifier, \texttt{+GPT-5}, which serves as an \emph{upper bound} for the edge-level clarification paradigm, helping to establish the role of \texttt{+GPT-5} as a ceiling reference rather than a competing baseline.
We assess whether a lightweight, \emph{edge-level} clarifier improves reliability under unchanged orchestration, how it trades accuracy against latency and cost, and why these effects arise. Our study is structured around four research questions (\textbf{RQ}):
\begin{itemize}[leftmargin=*, topsep=2pt, itemsep=2pt]
  \item \textbf{RQ1: Paradigm validity and end-to-end effectiveness.}
  Under unchanged orchestration, does inserting an edge-level clarifier consistently improve end-task performance across frameworks and benchmarks?
  \item \textbf{RQ2: Efficiency under practical budgets.}
  Can \texttt{AgentAsk} achieve these gains with low overhead in latency and extra cost, and how does it compare to the \texttt{+GPT-5} upper-bound clarifier in the accuracy--efficiency trade-off?
  \item \textbf{RQ3: Ablations and robustness.}
  Which components are necessary for the observed gains, and how stable is performance under key design choices and control parameters?
  \item \textbf{RQ4: Taxonomy-grounded mechanisms.}
Which error types drive most interventions and gains, and how does the edge-level clarification improve efficiency while minimizing unnecessary overhead?

\vspace{-0.1cm}
\end{itemize}
\subsection{Experimental Setup}
\paragraph{Benchmarks \& Baselines.} We evaluate \textbf{AgentAsk} on five public benchmarks spanning math reasoning (GSM8K \citep{gsm8k}, MATH \citep{MATH}), general-knowledge QA (MMLU \citep{mmlu}), and code generation (HumanEval \citep{humaneval}, MBPP \citep{mbpp}). Baselines include single-model prompting (IO, CoT, Self-Refine \citep{cot,selfrefine}) and four representative multi-agent frameworks (GPTSwarm \citep{gptswarm}, AFlow \citep{aflow}, MaAS \cite{maas}, MasRouter \citep{masrouter}).

\paragraph{Implementation Details.} For each framework, we consider three settings that isolate the effect of the clarifier while holding orchestration constant: (i) \texttt{origin} uses the public implementation; (ii) \texttt{+GPT-5} substitutes the clarifier with \texttt{GPT-5} \citep{OpenAI2025gpt5}; (iii) \texttt{+AgentAsk} plugs in our lightweight module. The LLM executor is \texttt{GPT-4o-mini-0718} \citep{OpenAI2024gpt4omini}; \texttt{+AgentAsk} uses \texttt{Qwen-3-4B} \citep{yang2025qwen3technicalreport} as the backbone. Accuracy for GSM8K, MATH, and MMLU is reported; Pass@1 for HumanEval and MBPP. Due to baseline performance differences across frameworks, we normalize latency and extra cost by setting the \texttt{origin} framework to 100, and compare the relative increases for each configuration. Prompts are provided in the Appendix~\ref{sec:Prompt}. To better demonstrate our plug-and-play capabilities, we trained AgentAsk on one framework and then gradually integrated it into other frameworks.

\subsection{RQ1: Effectiveness under Fixed Orchestration}
\textbf{Observation.} Edge-level clarification consistently improves end-task performance under unchanged orchestration, suggesting that many failures originate at inter-agent handoff boundaries and can be mitigated by clarifying exchanged messages.

\noindent\textbf{Results and analysis.} We compare our lightweight \texttt{AgentAsk} with a strong clarifier, \texttt{+GPT-5}, which serves as an \emph{upper bound} for the edge-level clarification paradigm. Table~\ref{tab:main} shows that adding an edge-level clarifier yields consistent gains across frameworks and benchmarks. For each framework, both \texttt{+AgentAsk} and \texttt{+GPT-5} improve performance over \texttt{origin} on all five datasets, and the average score increases by a clear margin. In particular, \texttt{+AgentAsk} improves the framework-level averages by about \textbf{+3.29} to \textbf{+3.45} points, while \texttt{+GPT-5} improves them by about \textbf{+4.40} to \textbf{+4.69} points. 
% The gains are broad across task types, including math, general knowledge, and code generation, indicating that edge-level issues are a common source of failure. This pattern suggests that many failures arise at handoff boundaries and can be mitigated by clarifying exchanged messages rather than changing the orchestration itself.
The gains are consistent across diverse task types, including math, general knowledge, and code generation, indicating that edge-level issues are a common source of failure that can be mitigated by clarifying exchanged messages rather than altering the orchestration.
As expected, \texttt{+GPT-5} produces the largest improvements, supporting its role as an approximate upper bound for the clarification paradigm under the same pipeline. Notably, \texttt{+AgentAsk} recovers a substantial portion of this upper-bound gain with a lightweight learned module. The remaining gap is primarily attributable to efficiency considerations, which we analyze in RQ2.

To verify that the improvements are not specific to one decoding configuration, we additionally evaluate the main settings under multiple decoding temperatures and evaluation seeds, with full robustness results reported in Appendix~\ref{app:rq1}.

\begin{table}[t]
\centering
\small
\renewcommand{\arraystretch}{1.12}
\setlength{\tabcolsep}{6pt}
\begin{tabular}{lccc}
\Xhline{1.2pt}
\rowcolor{CadetBlue!16}
 & \multicolumn{3}{c}{\textbf{GSM8K}} \\ \cline{2-4} 
\rowcolor{CadetBlue!16} 
\multirow{-2}{*}{\textbf{Settings}} & \textit{Accuracy} & \textit{Latency} & \textit{Extra Cost} \\
\hline\hline
origin
& 93.26 & 100 & 0.0 \\

+\texttt{GPT\textendash 4o\textendash mini}
& 94.62{\inc{(+1.36)}} & 118 & 16.0 \\
+\texttt{GPT\textendash 5}
& 95.10{\inc{(+1.84)}} & 129 & 34.0 \\
\cdashline{1-4}
\rowcolor{gray!10}
\multicolumn{4}{l}{\textsl{AgentAsk} (\texttt{Backbone: Llama-3.2-3B})} \\
% \rowcolor{gray!10}
\quad SFT
& 93.64{\inc{(+0.38)}} & 106 & 5.7 \\
% \rowcolor{gray!10}
\quad \textbf{(E\textendash GRPO)}
& \textbf{94.23}{\inc{(\textbf{+0.97})}} & \textbf{105} & \textbf{5.0} \\
\cdashline{1-4}
\rowcolor{gray!10}
\multicolumn{4}{l}{\textsl{AgentAsk} (\texttt{Backbone: Qwen\textendash 3\textendash 4B)}} \\
\quad SFT
& 93.99{\inc{(+0.73)}} & 105 & 5.3 \\
\quad \textbf{(E\textendash GRPO)}
& \textbf{94.86}{\inc{(\textbf{+1.60})}} & \textbf{108} & \textbf{4.9} \\
\Xhline{1.2pt}
\end{tabular}
\caption{Results of experiments on \textbf{different types or training stages of clarifier} with MasRouter@GSM8K.}
\vspace{-0.6cm}
\label{tab:masrouter_breakdown}
\end{table}

\begin{figure*}
    \centering
    \includegraphics[width=\linewidth]{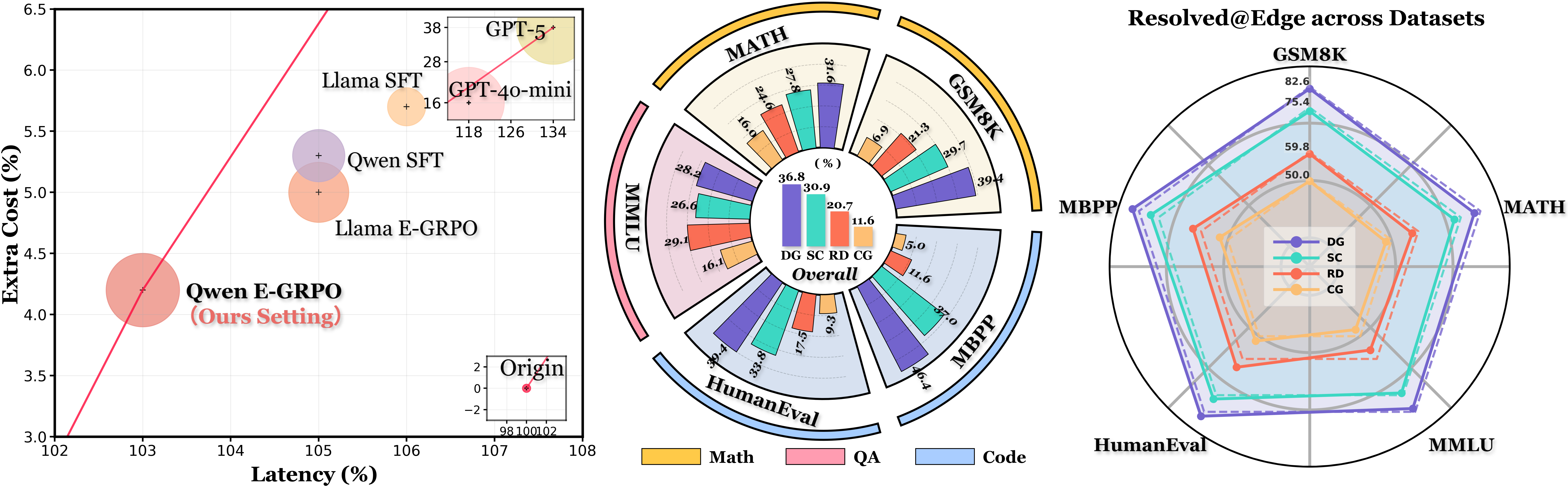}
    \caption{ \textbf{(Left)} Pareto frontier (\textit{x=latency, y=extra cost; bubble size=Acc}), showing \texttt{+AgentAsk} nearing \texttt{+GPT-5} at much lower overhead. \textbf{(Middle)} Error-Type distributions (DG/SC/RD/CG) across datasets. \textbf{(Right)}  Resolved@Edge of four error types across different datasets in AgentAsk.}
    \vspace{-0.2cm}
    \label{fig:exppic}
\end{figure*}

\subsection{RQ2: Efficiency under Practical Budgets}
% \label{sec:RQ2}
\textbf{Observation.} \texttt{AgentAsk} attains most of the improvement with lower latency and extra cost, making it more efficient than heavy clarifiers.

\noindent\textbf{Results and analysis.} We quantify the accuracy--efficiency trade-off on MasRouter@GSM8K in Table~\ref{tab:masrouter_breakdown}. Using \texttt{+GPT-5} yields the highest accuracy at \textbf{95.10}, but increases latency to \textbf{129} and extra cost to \textbf{34.0}. In contrast, \texttt{AgentAsk} with \texttt{Qwen-3-4B} achieves \textbf{94.86} accuracy with only \textbf{108} latency and \textbf{4.9} extra cost, reducing latency by \textbf{21} points and extra cost by \textbf{29.1} points while leaving a \textbf{0.24} point accuracy gap. The \texttt{Llama-3.2-3B} variant shows the same trend, further indicating that the efficiency advantage is not tied to a specific backbone.

Figure~\ref{fig:exppic} (Left) shows that \texttt{+AgentAsk} stays near the accuracy envelope with low overhead, while \texttt{+GPT-5} achieves slightly higher accuracy at much higher latency and cost. These results confirm that our edge-level clarifier is efficient under tight budgets, with the gap to the upper bound primarily due to capacity limitations rather than orchestration changes. We attribute this efficiency to the training objective that penalizes unnecessary interventions, encouraging the clarifier to ask only when the expected downstream benefit outweighs the overhead. Additional breakdowns across frameworks and datasets are reported in Appendix~\ref{app:rq2}.

\subsection{RQ3: Mechanism and Robustness at Edge}
\textbf{Observation.} The effectiveness of \texttt{AgentAsk} is driven by its ability to intervene at communication handoffs, improving accuracy while maintaining efficiency. Its robustness is confirmed through ablation studies and sensitivity analyses.

\noindent\textbf{Results and analysis.} Table~\ref{tab:masrouter_mbpp_breakdown} presents the results of ablation experiments on the reward components for \textbf{AgentAsk} with the MBPP benchmark. Removing individual reward components leads to a decrease in accuracy and an increase in latency and cost, showing that each component contributes to balancing efficiency and performance. The removal of the \(r^{par}\) increases the latency and extra cost without much improvement in accuracy, emphasizing the importance of minimizing unnecessary clarifications for maintaining efficiency.
% Finally, Figure~\ref{fig:exppic} (Right) illustrates the sensitivity of \textbf{AgentAsk} to the window size \(H\) and the parsimony weight \(\lambda_{\mathrm{sw}}\), showing that the clarifier is robust to parameter variations, with stable performance across a range of settings. This confirms that \textbf{AgentAsk} is both effective and adaptable.
% The full set of results for all frameworks, including additional details on ablation studies and experiments with PPO, GRPO, and \textbf{AgentAsk}, can be found in Appendix~\ref{app:rq3}. 
Finally, the full set of results, including further details on ablation studies and experiments with PPO, GRPO, and our method, can be found in Appendix~\ref{app:rq3}, where Figure~\ref{fig:sensitivity} demonstrates the sensitivity.
\begin{table}[t]
\centering
\fontsize{9}{11}\selectfont
\renewcommand{\arraystretch}{1.12}
\setlength{\tabcolsep}{8pt}
\begin{tabular}{lccc}
\Xhline{1.2pt}
\rowcolor{CadetBlue!16}
 & \multicolumn{3}{c}{\textbf{MBPP}} \\ \cline{2-4} 
\rowcolor{CadetBlue!16}
\multirow{-2}{*}{\textbf{Settings}}
& \textit{Accuracy} & \textit{Latency} & \textit{Extra Cost} \\
\hline\hline

\textbf{\textsl{AgentAsk}}
& 83.55 & 103 & 7.7 \\
\rowcolor{gray!10}
\textbf{w/o $r^{par}$}
& 81.34{\dec{(-2.21)}} & 118 & 9.6 \\
\textbf{w/o $r^{eff}$}
& 80.05{\dec{(-3.50)}} & 108 & 7.0 \\
\rowcolor{gray!10}
\textbf{w/o $r^{fmt}$}
& 81.42{\dec{(-2.13)}} & 111 & 9.7 \\
\textbf{$R$ only}
& 79.66{\dec{(-3.89)}} & 123 & 13.2 \\

\Xhline{1.2pt}
\end{tabular}
\caption{Results of ablation experiments on \textbf{different reward settings} with MasRouter@MBPP(red \textcolor{red!70!black}{$\downarrow$} represents the decrease to each framework’s \texttt{AgentAsk}).}
\vspace{-0.4cm}
\label{tab:masrouter_mbpp_breakdown}
\end{table}

\vspace{-0.1cm}
\subsection{RQ4: Taxonomy Mechanisms at Edge}
\textbf{Observation.} The performance of \texttt{AgentAsk} is influenced by the types of errors it addresses, with varying effectiveness across different error types.%The performance of \texttt{AgentAsk} depends on the types of errors it addresses, which we analyze across benchmarks.

\noindent\textbf{Results and analysis.} Figure~\ref{fig:exppic} (Middle and Right) shows the distribution of error types across datasets, along with the Resolved@Edge percentages for each error type. Resolved@Edge represents the percentage of clarifications that successfully resolve the error without further intervention (illustrated in Eq~\ref{eq:rl_eff}). AgentAsk performs particularly well in tasks dominated by Data Gap and Signal Corruption errors. These errors are typically resolved with a single clarification, leading to high Resolved@Edge rates, 82.5\% for DG and 75.4\% for SC. DG makes up 36.8\% and SC 30.9\% of the errors across datasets, representing the majority of the issues encountered. For tasks involving more Referential Drift and Capability Gap, such as MATH and MMLU, the improvements are more limited. These errors are less frequent, with RD accounting for 20.7\% and CG 11.6\% of the errors, but they are more difficult to resolve with a single clarification. As a result, their Resolved@Edge rates are lower, 58.3\% for RD and 49.5\% for CG.

Overall, AgentAsk is highly effective in resolving simpler edge-level errors. However, for tasks with more complex errors like RD and CG, the improvements are more modest, as these errors stems from the nature of the hallucinations. For further analysis and additional results including qualitative case studies, please refer to Appendix~\ref{app:case}.

% Figure~\ref{fig:exppic} (Middle) shows the distribution of error types across datasets. \texttt{AgentAsk} excels in tasks dominated by Data Gap and Signal Corruption errors, which are typically resolved with a single clarification, leading to significant accuracy gains with minimal latency and cost.
% However, for tasks involving Referential Drift and Capability Gap errors, such as MATH and MMLU, the gains are more limited. These errors often result from the model’s hallucinations, where incorrect assumptions or ambiguous references lead to propagated errors. While \texttt{AgentAsk} helps mitigate such issues at the edge, fully resolving these more complex errors may require deeper reasoning beyond a single clarification. This limitation stems from the nature of the hallucinations, rather than a flaw in the edge-level clarification mechanism itself.
% Overall, \texttt{AgentAsk} is highly effective in resolving simpler edge-level errors and provides substantial improvements in tasks where such errors are prevalent. The findings underscore the practicality of \texttt{AgentAsk} in systems with frequent edge-level issues, even if it faces challenges with hallucinations in more complex tasks.

% \vspace{-0.1cm}
\section{Conclusion}
\vspace{-0.1cm}
In this work, we provide novel insights into error propagation in multi-agent systems, emphasizing the critical role of communication at agent handoffs. Our error taxonomy identifies four common error types that drive system failures, laying the groundwork for effective intervention. Building on this taxonomy, we propose AgentAsk, a lightweight, architecture-agnostic clarifier that minimizes intervention by asking targeted questions at key communication points. This approach enables error containment with minimal overhead, achieving near large-model performance. Future work will explore uncertainty‑aware gating, theoretical analysis of edge‑level interventions, and integration with tools and human feedback.
\newpage

\section*{Limitations}

% AgentAsk currently inherits much of its judgment quality from the underlying LLMs used as the teacher (for distillation) and as the lightweight clarifier backbone. As a result, its reliability and granularity of clarifications scale with these base models; under weaker backbones, gains may diminish even though the logic of edge-level intervention remains sound. Future work will explore model-agnostic uncertainty signals and hybrid features to reduce dependence on specific LLM strengths.
Despite its effectiveness in mitigating many forms of error propagation, AgentAsk cannot fully eliminate all errors within multi-agent systems. Certain errors originate from internal flaws within individual agents, particularly those tied to well-known limitations in large models, such as hallucinations and inconsistencies. These issues lie beyond the scope of a clarification module like AgentAsk and can only be addressed through more fundamental advancements in the capabilities of large models. Future research will need to focus on overcoming these intrinsic model limitations, with the hope that improvements in model robustness and accuracy will enable more reliable error resolution across the system.

% \section*{Ethical Considerations}

% This work adheres to the ACL Code of Ethics through its use of publicly available benchmarks, which contain no personally identifiable information or sensitive content, ensuring all experiments are fully computational and involve no human participants. The proposed AgentAsk framework employs open-source language models as its backbone clarifiers, enhancing transparency in design and implementation. To promote reproducibility, we provide full implementation code and detailed experimental configurations.

\bibliography{acl_latex}
\newpage
\appendix

\tableofcontents

\section{Supplementary Design Explaination}
\subsection{Modeling Details}
\label{app:controller_details}

We elaborate on the controller framework referenced in Section~\ref{sec:modeling}.

\paragraph{Edge State.} Each input produces a message graph $G = (V, E)$ unrolled into an edge sequence $\{e_t\}_{t=1}^{T}$, where each $e_t = (u_t \rightarrow v_t)$ denotes a message handoff. The controller observes a local state at each step:
\begin{equation}
x_t = \left(x^{\mathrm{in}}_t,\; u_t,\; v_t,\; m_t,\; h_t\right),
\end{equation}
where $x^{\mathrm{in}}_t$ is the user query or subgoal, $m_t$ is the message from $u_t$, and $h_t$ encodes recent local interaction context.

\paragraph{Controller Action.} Based on $x_t$, the controller selects an action:
\begin{equation}
a_t = (z_t, \tilde{v}_t, q_t),
\end{equation}
where $z_t$ is a binary gate indicating whether to ask, $\tilde{v}_t$ selects which agent to query, and $q_t$ is the generated clarification question.

\paragraph{Policy Structure.} We define the controller policy as:
\begin{equation}
\label{eq:policy}
\pi_\theta(a_t \mid x_t) = \pi_\theta(z_t \mid x_t)\; \left[\pi_\theta(\tilde{v}_t \mid x_t)\; \pi_\theta(q_t \mid x_t)\right]^{z_t}.
\end{equation}

\paragraph{Trajectory Modeling.} The full input interaction produces a trajectory:
\begin{equation}
\tau = \left\{(x_t, a_t, r_t)\right\}_{t=1}^{T},
\end{equation}
where $r_t$ is a response returned when $z_t = 1$. The objective from Section~\ref{sec:modeling} is instantiated here with:
\begin{itemize}
  \item $U(\tau)$ rewarding task success and helpful clarifications,
  \item $C(\tau)$ measuring token and latency costs,
  \item $B$ as the intervention budget.
\end{itemize}

This setup supports both imitation-based initialization and constrained reinforcement learning (see Appendix~\ref{app:sft_loss}).

\begin{algorithm*}[!t]
\caption{AgentAsk Training}
\label{alg:agentask}
\begin{algorithmic}[1]
\Require Logs $\mathcal{L}$, teacher $\mathcal{J}$, init policy $\pi_\theta$, ref $\pi_{\mathrm{ref}}$, budget $B$, lr $\eta$, $\lambda_{\text{ask}}$, $\lambda_R$, $\epsilon$, $\beta$, weights $w_t$, baseline $b$

\State \underline{\textbf{Stage A: Build  Corpus}}
\For{trajectory $\tau \in \mathcal{L}$}
  \For{edge state $x_i \in \tau$}
    \State $y_i \gets \mathcal{J}(x_i)=(t_i,v_i,q_i)$ \textcolor{gray}{\Comment{$t_i\in\{\texttt{DG},\texttt{SC},\texttt{RD},\texttt{CG},\texttt{NONE}\}$}}
    \State $\widetilde{\mathcal{D}} \gets \widetilde{\mathcal{D}} \cup \{(x_i,y_i)\}$
  \EndFor
\EndFor

\State \underline{\textbf{Stage B: Learning from Demonstrations}}
\Repeat
  \State Sample minibatch $\mathcal{B}\subset\widetilde{\mathcal{D}}$
  \State Compute heads $p_\theta^{\text{type}}(t\!\mid\!x)$, $p_\theta^{\text{addr}}(v\!\mid\!x,t)$, $p_\theta^{\text{txt}}(q\!\mid\!x,t)$
  \State Evaluate $\mathcal{L}_{\text{type}}$ and $\mathcal{L}_{\text{ask}}$; set $\mathcal{L}_{\text{Total}}=\mathcal{L}_{\text{type}}+\lambda_{\text{ask}}\mathcal{L}_{\text{ask}}$
  \State $\theta \gets \theta - \eta\nabla_\theta \mathcal{L}_{\text{Total}}$
\Until{validation stops improving}

\State \underline{\textbf{Stage C:Clarification Optimization with RL}}
\Repeat
  \State Init episode, counters, buffers; set budget $B$
  \For{$t=1,\ldots,T$}
    \State Observe $x_t=(x_t^{\mathrm{in}},u_t,v_t,m_t,h_t)$
    \State Sample $a_t=(z_t,\tilde v_t,q_t)\sim \pi_\theta(\cdot\mid x_t)$ (schema+budget)
    \State Execute $a_t$; if $z_t{=}1$ get reply $r_t$; deliver to $v_t$; obtain $x_{t+1}$
    \State Compute $r_t^{\mathrm{eff}},r_t^{\mathrm{par}},r_t^{\mathrm{fmt}},r_t^{\mathrm{edge}}$
    \If{episode not terminated}
       \State Sample $x_t$; score by $r_t^{\mathrm{edge}}$
       \State Form local advantages $A_{t}^{\mathrm{loc}}$; update with ratios $\rho_{t}$ 
    \Else
       \State Compute terminal $R$; set $A_{t'}^{\mathrm{glob}} \gets w_{t'}(R-b)$ for all visited $t'$
       \State Apply E\text{-}GRPO update using $A_{t'}^{\mathrm{loc}}$ and $A_{t'}^{\mathrm{glob}}$ with KL to $\pi_{\mathrm{ref}}$
       \State \textbf{break}
    \EndIf
  \EndFor
\Until{RL budget exhausted or convergence}
\State \Return $\pi_\theta$
\end{algorithmic}
\end{algorithm*}

\subsection{Supervised Fine-tuning Details}
\label{app:sft_loss}

We detail the data construction and training procedure used in Section~4.2.

\paragraph{Dataset Construction.} From a set of logged multi-agent executions, we extract edge-local contexts $x_i$ and assign expert clarifications via a teacher model. Each label $y_i = (t_i, v_i, q_i)$ encodes:
\begin{itemize}
  \item $t_i \in \mathcal{T}$, a clarification type (or $\texttt{NONE}$ if no ask is needed),
  \item $v_i \in \{u_i, v_i\}$, the agent selected to receive the question,
  \item $q_i$, the clarification question string, constrained by schema and token length.
\end{itemize}

We define $m_i = \mathbbm{1}[t_i \ne \texttt{NONE}]$ as an indicator for whether a clarification is present.

\paragraph{Model Outputs.} The controller predicts:
\begin{itemize}
  \item $p^{\text{type}}_\theta(t_i \mid x_i)$ for the ask decision,
  \item $p^{\text{addr}}_\theta(v_i \mid x_i, t_i)$ for the addressee (when asking),
  \item $p^{\text{txt}}_\theta(q_i \mid x_i, t_i)$ for the clarification text via auto-regressive decoding.
\end{itemize}

\paragraph{Training Objective.} The supervised loss has two components:
\begin{itemize}
  \item $\mathcal{L}_{\text{type}}$ is the cross-entropy loss on $t_i$,
  \item $\mathcal{L}_{\text{ask}}$ is the conditional loss on $v_i$ and $q_i$, applied only when $m_i=1$:
\end{itemize}
\begin{equation}
\label{eq:sft_ask}
\resizebox{\columnwidth}{!}{$
\begin{aligned}
\mathcal{L}_{\text{ask}}
=\frac{1}{N}\sum_{i=1}^{N} m_i\Big[&-\log p_\theta^{\text{addr}}(v_i\mid x_i,t_i)
\\
&-\sum_{t=1}^{T_i}\log p_\theta^{\text{txt}}(q_{i,t}\mid x_i,t_i,q_{i,<t})\Big].
\end{aligned}
$}
\end{equation}

The total training loss is illustrated at Equation~\ref{eq:sft_total}
where $\lambda_{\text{ask}}$ controls the relative weight of question generation.

This training yields a controller capable of structured clarification grounded in edge-level context.

\begin{table*}[!t]
\centering
\small
\renewcommand{\arraystretch}{1.2}
\setlength{\tabcolsep}{6pt}
\begin{tabular}{llcccccc}
\Xhline{1.2pt}
\rowcolor{CadetBlue!16}
\textbf{Method} &\textbf{Settings}  & \textbf{GSM8K} & \textbf{MATH} & \textbf{HumanEval} & \textbf{MMLU} & \textbf{MBPP} & \textbf{Average} \\
\hline \hline
\rowcolor{gray!10}
IO & & 89.52 & 49.23 & 89.66 & 79.63 & 71.38 & 75.88 \\
CoT & & 89.71 & 48.19 & 90.64 & 80.44 & 70.66 & 75.93 \\
\rowcolor{gray!10}
Self-Refine & & 90.37 & 47.92 & 88.88 & 78.31 & 71.49 & 75.40 \\
\hline

\multirow{3}{*}{GPTSwarm}
  & origin       & 91.82 & 50.92 & 90.21 & 80.92 & 76.38 & 77.99 \\
  & \texttt{+GPT-5}         & \rbdelta{95.35}{\inc{+3.53}} & \rbdelta{56.15}{\inc{+5.23}} & \rbdelta{93.35}{\inc{+3.14}} & \rbdelta{86.20}{\inc{+5.28}} & \rbdelta{82.05}{\inc{+5.67}} & 82.62\inc{+4.63} \\
  & +AgentAsk    & \rbdelta{94.17}{\inc{+2.35}} & \rbdelta{54.97}{\inc{+4.05}} & \rbdelta{92.41}{\inc{+2.20}} & \rbdelta{84.62}{\inc{+3.70}} & \rbdelta{80.48}{\inc{+4.10}} & 81.33\inc{+3.34} \\
\hline

\rowcolor{gray!10}
  & origin       & 91.02 & 50.51 & 91.22 & 81.60 & 79.14 & 78.70 \\
\rowcolor{gray!10}
  \multirow{-1}{*}{AFlow}
  & \texttt{+GPT-5}         & \rbdelta{94.85}{\inc{+3.83}} & \rbdelta{56.05}{\inc{+5.54}} & \rbdelta{94.60}{\inc{+3.38}} & \rbdelta{86.65}{\inc{+5.05}} & \rbdelta{84.75}{\inc{+5.61}} & 83.38\inc{+4.68} \\
\rowcolor{gray!10}
  & +AgentAsk    & \rbdelta{93.12}{\inc{+2.10}} & \rbdelta{54.71}{\inc{+4.20}} & \rbdelta{93.57}{\inc{+2.35}} & \rbdelta{85.25}{\inc{+3.65}} & \rbdelta{83.04}{\inc{+3.90}} & 81.94\inc{+3.24} \\
\hline

\multirow{3}{*}{MaAS}
  & origin       & 92.32 & 50.82 & 91.42 & 82.41 & 80.80 & 79.75 \\
  & \texttt{+GPT-5}         & \rbdelta{95.40}{\inc{+3.08}} & \rbdelta{56.70}{\inc{+5.88}} & \rbdelta{94.85}{\inc{+3.43}} & \rbdelta{87.10}{\inc{+4.69}} & \rbdelta{86.20}{\inc{+5.40}} & 84.05\inc{+4.30} \\
  & +AgentAsk    & \rbdelta{92.12}{\dec{-0.20}} & \rbdelta{55.42}{\inc{+4.60}} & \rbdelta{93.67}{\inc{+2.25}} & \rbdelta{85.81}{\inc{+3.40}} & \rbdelta{84.55}{\inc{+3.75}} & 82.31\inc{+2.56} \\
\hline

\rowcolor{gray!10}
  & origin       & 92.82 & 51.10 & 90.72 & 82.60 & 78.58 & 79.16 \\
\rowcolor{gray!10}
  \multirow{-1}{*}{MasRouter}
  & \texttt{+GPT-5}         & \rbdelta{96.00}{\inc{+3.18}} & \rbdelta{56.25}{\inc{+5.15}} & \rbdelta{93.95}{\inc{+3.23}} & \rbdelta{86.90}{\inc{+4.30}} & \rbdelta{84.60}{\inc{+6.02}} & 83.54\inc{+4.38} \\
\rowcolor{gray!10}
  & +AgentAsk    & \rbdelta{95.07}{\inc{+2.25}} & \rbdelta{54.95}{\inc{+3.85}} & \rbdelta{92.82}{\inc{+2.10}} & \rbdelta{85.70}{\inc{+3.10}} & \rbdelta{82.83}{\inc{+4.25}} & 82.27\inc{+3.11} \\
\Xhline{1.2pt}
\end{tabular}
\caption{Accuracy/Pass@1 across frameworks under three settings (\texttt{origin}, \texttt{+GPT-5}, \texttt{+AgentAsk}) on five benchmarks. Bottom-right overlays denote percentage-point deltas vs.\ each framework’s \texttt{origin}. Each cell overlays a bottom-right delta (green \textcolor{green!60!black}{$\uparrow$} / red \textcolor{red!70!black}{$\downarrow$}) relative to the framework’s \texttt{origin}.}
\vspace{-0.2cm}
\label{tab:main_appendix}
\end{table*}

\begin{table*}[!t]
\centering
\small
\renewcommand{\arraystretch}{1.12}
\setlength{\tabcolsep}{4.2pt}
\resizebox{\textwidth}{!}{%
\begin{tabular}{l|ccc|ccc|ccc|ccc|ccc}
\Xhline{1.2pt}
\rowcolor{CadetBlue!16}
\multirow{1}{*}{\textbf{Settings}} &
\multicolumn{3}{c|}{\textbf{GSM8K}} &
\multicolumn{3}{c|}{\textbf{MATH}} &
\multicolumn{3}{c|}{\textbf{HumanEval}} &
\multicolumn{3}{c|}{\textbf{MMLU}} &
\multicolumn{3}{c}{\textbf{MBPP}} \\
\rowcolor{CadetBlue!16}
& Acc. & Lat. & Extra & Acc. & Lat. & Extra & Acc. & Lat. & Extra & Acc. & Lat. & Extra & Acc. & Lat. & Extra \\
\Xhline{1.2pt}
origin
& 92.18 & 100 & 0.0 & 51.41 & 100 & 0.0 & 90.63 & 100 & 0.0 & 81.26 & 100 & 0.0 & 76.81 & 100 & 0.0 \\
+GPT\textendash 4o\textendash mini
& \rbdelta{93.50}{\inc{+1.32}} & 117 & 16.0 & \rbdelta{51.84}{\inc{+0.43}} & 115 & 14.8 & \rbdelta{91.92}{\inc{+1.29}} & 121 & 17.6 & \rbdelta{82.00}{\inc{+0.74}} & 114 & 14.2 & \rbdelta{78.30}{\inc{+1.49}} & 123 & 18.5 \\
+GPT\textendash 5
& \rbdelta{95.55}{\inc{+3.37}} & 133 & 37.2 & \rbdelta{56.30}{\inc{+4.89}} & 131 & 35.8 & \rbdelta{94.05}{\inc{+3.42}} & 139 & 40.4 & \rbdelta{86.85}{\inc{+5.59}} & 129 & 33.8 & \rbdelta{82.45}{\inc{+5.64}} & 140 & 41.2 \\
\cdashline{1-16}
\rowcolor{gray!10}
\multicolumn{16}{l}{\textsl{AgentAsk} (\texttt{Backbone: Llama-3.2-3B})} \\
% \rowcolor{gray!10}
\quad SFT
& \rbdelta{92.60}{\inc{+0.42}} & 106 & 5.6 & \rbdelta{51.55}{\inc{+0.14}} & 106 & 5.4 & \rbdelta{91.25}{\inc{+0.62}} & 106 & 5.6 & \rbdelta{81.65}{\inc{+0.39}} & 105 & 5.0 & \rbdelta{76.70}{\dec{-0.11}} & 107 & 5.7 \\
% \rowcolor{gray!10}
\quad \textbf{(E\textendash GRPO)}
& \rbdelta{92.88}{\inc{+0.70}} & 105 & 5.0 & \rbdelta{51.65}{\inc{+0.24}} & 105 & 5.1 & \rbdelta{91.50}{\inc{+0.87}} & 105 & 5.2 & \rbdelta{81.90}{\inc{+0.64}} & 104 & 4.7 & \rbdelta{77.05}{\inc{+0.24}} & 106 & 5.2 \\
\cdashline{1-16}
\rowcolor{gray!10}
\multicolumn{16}{l}{\textsl{AgentAsk} (\texttt{Backbone: Qwen\textendash 3\textendash 4B)}} \\
\quad SFT
& \rbdelta{92.85}{\inc{+0.67}} & 106 & 5.5 & \rbdelta{51.41}{\ \ \ \ 0.00} & 106 & 5.3 & \rbdelta{91.40}{\inc{+0.77}} & 105 & 5.3 & \rbdelta{81.80}{\inc{+0.54}} & 105 & 5.0 & \rbdelta{77.05}{\inc{+0.24}} & 106 & 5.5 \\
\quad \textbf{(E\textendash GRPO)}
& \rbdelta{94.52}{\inc{+2.34}} & 116 & 9.7 & \rbdelta{55.05}{\inc{+3.64}} & 106 & 9.2 & \rbdelta{92.95}{\inc{+2.32}} & 103 & 3.3 & \rbdelta{85.25}{\inc{+3.99}} & 119 & 8.6 & \rbdelta{81.15}{\inc{+4.34}} & 105 & 6.1 \\
\Xhline{1.2pt}
\end{tabular}
}%
\caption{\textbf{GPTSwarm: Ablations on model and training across five benchmarks.}}
\label{tab:gptswarm_all}
\end{table*}

\begin{table*}[!t]
\centering
\small
\renewcommand{\arraystretch}{1.12}
\setlength{\tabcolsep}{4.2pt}
\resizebox{\textwidth}{!}{%
\begin{tabular}{l|ccc|ccc|ccc|ccc|ccc}
\Xhline{1.2pt}
\rowcolor{CadetBlue!16}
\multirow{1}{*}{\textbf{Settings}} &
\multicolumn{3}{c|}{\textbf{GSM8K}} &
\multicolumn{3}{c|}{\textbf{MATH}} &
\multicolumn{3}{c|}{\textbf{HumanEval}} &
\multicolumn{3}{c|}{\textbf{MMLU}} &
\multicolumn{3}{c}{\textbf{MBPP}} \\
\rowcolor{CadetBlue!16}
& Acc. & Lat. & Extra & Acc. & Lat. & Extra & Acc. & Lat. & Extra & Acc. & Lat. & Extra & Acc. & Lat. & Extra \\
\Xhline{1.2pt}
origin
& 92.84 & 100 & 0.0 & 51.08 & 100 & 0.0 & 92.02 & 100 & 0.0 & 82.93 & 100 & 0.0 & 81.22 & 100 & 0.0 \\
+GPT\textendash 4o\textendash mini
& \rbdelta{93.95}{\inc{+1.11}} & 118 & 15.7 & \rbdelta{51.80}{\inc{+0.72}} & 116 & 14.9 & \rbdelta{92.80}{\inc{+0.78}} & 120 & 16.6 & \rbdelta{83.40}{\inc{+0.47}} & 115 & 14.3 & \rbdelta{82.10}{\inc{+0.88}} & 122 & 17.3 \\
+GPT\textendash 5
& \rbdelta{96.20}{\inc{+3.36}} & 127 & 44.3 & \rbdelta{56.85}{\inc{+5.77}} & 132 & 27.7 & \rbdelta{95.45}{\inc{+3.43}} & 121 & 17.8 & \rbdelta{88.35}{\inc{+5.42}} & 130 & 31.0 & \rbdelta{86.70}{\inc{+5.48}} & 129 & 35.6 \\
\cdashline{1-16}
\rowcolor{gray!10}
\multicolumn{16}{l}{\textsl{AgentAsk} (\texttt{Backbone: Llama-3.2-3B})} \\
% \rowcolor{gray!10}
\quad SFT
& \rbdelta{92.90}{\inc{+0.06}} & 106 & 5.4 & \rbdelta{51.35}{\inc{+0.27}} & 106 & 5.2 & \rbdelta{91.90}{\dec{-0.12}} & 107 & 5.6 & \rbdelta{83.05}{\inc{+0.12}} & 105 & 4.9 & \rbdelta{81.60}{\inc{+0.38}} & 107 & 5.6 \\
% \rowcolor{gray!10}
\quad \textbf{(E\textendash GRPO)}
& \rbdelta{93.00}{\inc{+0.16}} & 105 & 5.0 & \rbdelta{51.60}{\inc{+0.52}} & 105 & 5.0 & \rbdelta{92.20}{\inc{+0.18}} & 106 & 5.3 & \rbdelta{83.18}{\inc{+0.25}} & 104 & 4.7 & \rbdelta{81.85}{\inc{+0.63}} & 106 & 5.3 \\
\cdashline{1-16}
\rowcolor{gray!10}
\multicolumn{16}{l}{\textsl{AgentAsk} (\texttt{Backbone: Qwen\textendash 3\textendash 4B)}} \\
\quad SFT
& \rbdelta{92.84}{\ \ \ \ 0.00} & 106 & 5.4 & \rbdelta{51.90}{\inc{+0.82}} & 106 & 5.3 & \rbdelta{92.10}{\inc{+0.08}} & 106 & 5.5 & \rbdelta{82.85}{\dec{-0.08}} & 105 & 5.0 & \rbdelta{81.70}{\inc{+0.48}} & 106 & 5.4 \\
\quad \textbf{(E\textendash GRPO)}
& \rbdelta{95.10}{\inc{+2.26}} & 108 & 5.6 & \rbdelta{55.55}{\inc{+4.47}} & 112 & 7.4 & \rbdelta{94.45}{\inc{+2.43}} & 104 & 5.3 & \rbdelta{86.95}{\inc{+4.02}} & 117 & 5.7 & \rbdelta{85.30}{\inc{+4.08}} & 107 & 5.0 \\
\Xhline{1.2pt}
\end{tabular}
}%
\caption{\textbf{MaAS: Ablations on model and training across five benchmarks.}}
\label{tab:maas_all}
\end{table*}

\begin{table*}[!t]
\centering
\small
\renewcommand{\arraystretch}{1.12}
\setlength{\tabcolsep}{4.2pt}
\resizebox{\textwidth}{!}{%
\begin{tabular}{l|ccc|ccc|ccc|ccc|ccc}
\Xhline{1.2pt}
\rowcolor{CadetBlue!16}
 &
\multicolumn{3}{c|}{\textbf{GSM8K}} &
\multicolumn{3}{c|}{\textbf{MATH}} &
\multicolumn{3}{c|}{\textbf{HumanEval}} &
\multicolumn{3}{c|}{\textbf{MMLU}} &
\multicolumn{3}{c}{\textbf{MBPP}} \\
\rowcolor{CadetBlue!16}
\multirow{-2}{*}{\textbf{Settings}}
& Acc. & Lat. & Extra & Acc. & Lat. & Extra & Acc. & Lat. & Extra & Acc. & Lat. & Extra & Acc. & Lat. & Extra \\
\Xhline{1.2pt}
origin
& 93.26 & 100 & 0.0 & 51.52 & 100 & 0.0 & 91.03 & 100 & 0.0 & 83.19 & 100 & 0.0 & 79.04 & 100 & 0.0 \\
+GPT\textendash 4o\textendash mini
& \rbdelta{94.62}{\inc{+1.36}} & 118 & 16.0 & \rbdelta{52.10}{\inc{+0.58}} & 118 & 15.7 & \rbdelta{92.10}{\inc{+1.07}} & 122 & 17.3 & \rbdelta{83.50}{\inc{+0.31}} & 116 & 14.6 & \rbdelta{81.10}{\inc{+2.06}} & 121 & 17.9 \\
+GPT\textendash 5
& \rbdelta{95.10}{\inc{+1.84}} & 129 & 34.0 & \rbdelta{56.70}{\inc{+5.18}} & 124 & 24.8 & \rbdelta{94.55}{\inc{+3.52}} & 135 & 35.1 & \rbdelta{88.55}{\inc{+5.36}} & 128 & 29.3 & \rbdelta{85.15}{\inc{+6.11}} & 134 & 42.6 \\
\cdashline{1-16}
\rowcolor{gray!10}
\multicolumn{16}{l}{\textsl{AgentAsk} (\texttt{Backbone: Llama-3.2-3B})} \\
% \rowcolor{gray!10}
\quad SFT
& \rbdelta{93.64}{\inc{+0.38}} & 106 & 5.7 & \rbdelta{51.85}{\inc{+0.33}} & 107 & 5.8 & \rbdelta{91.35}{\inc{+0.32}} & 106 & 5.6 & \rbdelta{83.25}{\inc{+0.06}} & 106 & 5.2 & \rbdelta{79.80}{\inc{+0.76}} & 106 & 5.9 \\
% \rowcolor{gray!10}
\quad \textbf{(E\textendash GRPO)}
& \rbdelta{94.23}{\inc{+0.97}} & 105 & 5.0 & \rbdelta{51.95}{\inc{+0.43}} & 106 & 5.4 & \rbdelta{91.50}{\inc{+0.47}} & 105 & 5.2 & \rbdelta{83.35}{\inc{+0.16}} & 105 & 5.0 & \rbdelta{80.20}{\inc{+1.16}} & 105 & 5.4 \\
\cdashline{1-16}
\rowcolor{gray!10}
\multicolumn{16}{l}{\textsl{AgentAsk} (\texttt{Backbone: Qwen\textendash 3\textendash 4B)}} \\
\quad SFT
& \rbdelta{93.99}{\inc{+0.73}} & 105 & 5.3 & \rbdelta{51.90}{\inc{+0.38}} & 106 & 5.2 & \rbdelta{91.45}{\inc{+0.42}} & 105 & 5.1 & \rbdelta{83.10}{\dec{-0.09}} & 104 & 4.8 & \rbdelta{80.10}{\inc{+1.06}} & 104 & 4.9 \\
\quad \textbf{(E\textendash GRPO)}
& \rbdelta{94.86}{\inc{+1.60}} & 108 & 4.9 & \rbdelta{55.65}{\inc{+4.13}} & 109 & 6.1 & \rbdelta{93.35}{\inc{+0.52}} & 106 & 4.6 & \rbdelta{87.10}{\inc{+2.32}} & 111 & 5.7 & \rbdelta{83.55}{\inc{+4.51}} & 103 & 7.7 \\
\Xhline{1.2pt}
\end{tabular}
}%
\caption{\textbf{MasRouter: Ablations on model and training across five benchmarks.} }
\label{tab:masrouter_all}
\end{table*}

\begin{table*}[!t]
\centering
\small
\renewcommand{\arraystretch}{1.2}
\setlength{\tabcolsep}{2.5pt}
\begin{tabular}{llccc ccc ccc ccc ccc}
\Xhline{1.2pt}
\rowcolor{CadetBlue!16}
 & &
\multicolumn{3}{c}{\textbf{GSM8K}} &
\multicolumn{3}{c}{\textbf{MATH}} &
\multicolumn{3}{c}{\textbf{HumanEval}} &
\multicolumn{3}{c}{\textbf{MMLU}} &
\multicolumn{3}{c}{\textbf{MBPP}} \\ \cline{3-17}
\rowcolor{CadetBlue!16} \multirow{-2}{*}{\textbf{Framework}} & \multirow{-2}{*}{\textbf{Setting}}
& \textbf{Acc.} & \textbf{Lat.} & \textbf{Extra}
  & \textbf{Acc.} & \textbf{Lat.} & \textbf{Extra}
  & \textbf{Acc.} & \textbf{Lat.} & \textbf{Extra}
  & \textbf{Acc.} & \textbf{Lat.} & \textbf{Extra}
  & \textbf{Acc.} & \textbf{Lat.} & \textbf{Extra} \\
\hline\hline

\multirow{5}{*}{GPTSwarm}
& Ours              & 94.52 & 116 & 9.7 & 55.05 & 106 & 9.2 & 92.95 & 103 & 3.3 & 85.25 & 119 & 8.6 & 81.15 & 105 & 6.1 \\
& w/o $r^{par}$     & 93.10 & 126 & 6.2 & 52.38 & 116 & 7.0 & 91.32 & 113 & 7.0 & 82.33 & 118 & 6.5 & 78.07 & 115 & 8.3 \\
& w/o $r^{eff}$     & 92.84 & 117 & 6.8 & 52.75 & 125 & 7.6 & 91.85 & 109 & 5.9 & 83.02 & 117 & 7.3 & 82.14 & 109 & 7.2 \\
& w/o $r^{fmt}$     & 92.96 & 133 & 8.7 & 52.01 & 134 & 7.8 & 91.02 & 115 & 7.1 & 82.14 & 121 & 7.2 & 78.44 & 116 & 7.0 \\
& $R$ only            & 92.05 & 129 & 71.2 & 51.90 & 133 & 16.5 & 90.82 & 125 & 16.4 & 81.74 & 125 & 13.5 & 77.33 & 123 & 12.1 \\
\hline

\rowcolor{gray!10}
& Ours              & 94.86 & 108 & 4.9 & 55.65 & 109 & 6.1 & 93.35 & 106 & 4.6 & 87.10 & 111 & 5.7 & 83.55 & 103 & 7.7 \\
\rowcolor{gray!10}
& w/o $r^{par}$     & 94.61 & 113 & 6.4 & 52.98 & 117 & 7.2 & 92.48 & 112 & 6.7 & 83.74 & 116 & 6.9 & 81.34 & 118 & 9.6 \\
\rowcolor{gray!10}
\multirow{-1}{*}{MasRouter}

& w/o $r^{eff}$     & 94.70 & 111 & 6.9 & 53.22 & 115 & 7.4 & 92.91 & 109 & 5.9 & 83.88 & 113 & 6.3 & 80.05 & 108 & 7.0 \\
\rowcolor{gray!10}
& w/o $r^{fmt}$     & 94.51 & 129 & 29.8 & 52.44 & 138 & 27.7 & 92.15 & 119 & 16.8 & 82.90 & 128 & 17.0 & 80.42 & 111 & 9.7 \\
\rowcolor{gray!10}
& $R$ only            & 93.21 & 124 & 24.0 & 52.14 & 135 & 81.6 & 91.54 & 125 & 48.3 & 83.01 & 124 & 29.3 & 79.66 & 123 & 13.2 \\
\hline

\multirow{5}{*}{MaAS}
& Ours              & 95.10 & 108 & 5.6 & 55.55 & 112 & 7.4 & 94.45 & 104 & 5.3 & 86.95 & 108 & 5.7 & 85.30 & 107 & 5.0 \\
& w/o $r^{par}$     & 93.41 & 113 & 6.3 & 52.06 & 117 & 6.8 & 92.10 & 110 & 6.2 & 83.20 & 116 & 6.4 & 81.55 & 112 & 6.3 \\
& w/o $r^{eff}$     & 93.12 & 110 & 6.8 & 52.55 & 115 & 7.1 & 92.37 & 108 & 5.8 & 83.02 & 114 & 6.7 & 81.40 & 109 & 6.0 \\
& w/o $r^{fmt}$     & 92.98 & 114 & 7.4 & 51.74 & 118 & 7.6 & 91.61 & 112 & 6.6 & 82.60 & 118 & 7.0 & 80.44 & 113 & 6.8 \\
& $R$ only            & 92.55 & 125 & 25.3 & 51.60 & 128 & 14.4 & 92.14 & 121 & 13.6 & 83.02 & 114 & 9.4 & 81.40 & 119 & 13.6 \\

\Xhline{1.2pt}
\end{tabular}
\caption{Ablation results (Acc./Latency/Extra) across five benchmarks.}
\vspace{-0.2cm}
\label{tab:ablation_latency_extra}
\end{table*}

\begin{table*}[!t]
\centering
\small
\renewcommand{\arraystretch}{1.2}
\setlength{\tabcolsep}{6pt}
\begin{tabular}{llcccccc}
\Xhline{1.2pt}
\rowcolor{CadetBlue!16}
\textbf{Method} & \textbf{Settings} & \textbf{GSM8K} & \textbf{MATH} & \textbf{HumanEval} & \textbf{MMLU} & \textbf{MBPP} & \textbf{Average} \\
\hline\hline

\multirow{3}{*}{GPTSwarm}
 & PPO  & 92.18 & 51.41 & 90.63 & 81.26 & 76.81 & 78.46 \\
 & GRPO & 93.10 & 52.40 & 91.80 & 82.40 & 79.10 & 79.76 \\
 & Ours & 94.52 & 55.05 & 92.95 & 85.25 & 81.15 & 81.78 \\
\hline

\rowcolor{gray!10}
 & PPO  & 91.35 & 50.89 & 91.67 & 82.16 & 79.63 & 79.14 \\
\rowcolor{gray!10}
\multirow{-1}{*}{AFlow}
 & GRPO & 93.00 & 52.20 & 92.50 & 83.00 & 81.10 & 80.36 \\
\rowcolor{gray!10}
 & Ours & 93.70 & 54.95 & 93.95 & 86.40 & 83.90 & 82.58 \\
\hline

\multirow{3}{*}{MaAS}
 & PPO  & 92.84 & 51.08 & 92.02 & 82.93 & 81.22 & 80.02 \\
 & GRPO & 93.80 & 52.40 & 92.90 & 83.80 & 82.10 & 81.00 \\
 & Ours & 95.10 & 55.55 & 94.45 & 86.95 & 85.30 & 83.47 \\
\hline

\rowcolor{gray!10}
 & PPO  & 93.26 & 51.52 & 91.03 & 83.19 & 79.04 & 79.61 \\
\rowcolor{gray!10}
\multirow{-1}{*}{MasRouter}
 & GRPO & 94.70 & 52.80 & 92.70 & 84.10 & 81.90 & 81.24 \\
\rowcolor{gray!10}
 & Ours & 94.86 & 55.65 & 93.35 & 87.10 & 83.55 & 82.54 \\

\Xhline{1.2pt}
\end{tabular}
\caption{Performance comparison across PPO, GRPO, and our method on five benchmarks. }
\vspace{-0.2cm}
\label{tab:ppo_grpo_ours}
\end{table*}

\section{Algorithm Workflow}
\label{app:A}
This appendix concisely instantiates the edge controller in Eq.~\eqref{eq:policy} under the constraint in Eq.~\eqref{eq:constrained_obj}. The details can be seen in Section~\ref{sec:Methodology}. The specific algorithm process is shown in Algorithm~\ref{alg:agentask}.

\section{Supplementary Experiments}
\label{sec:Supplementary experiments}

\subsection{RQ1: Effectiveness under Fixed Orchestration}
\label{app:rq1}
As shown in Table~\ref{tab:main_appendix}, we present additional results evaluating the performance of \texttt{AgentAsk} across all frameworks and benchmarks under a random seed of 1314 and a model temperature of 0.3 settings. These results reinforce the trends observed in the main text, showing that \texttt{AgentAsk} consistently improves performance, particularly in terms of accuracy, latency, and extra cost. The accuracy improvement across the benchmarks, including GSM8K, MATH, HumanEval, MMLU, and MBPP, is substantial, with \texttt{AgentAsk} outperforming the baseline methods in all cases. For example, \texttt{AgentAsk} improves the average accuracy by +3.34\% on the GPTSwarm framework and achieves a similar performance boost on other benchmarks, while keeping latency and extra costs within acceptable bounds. These findings validate the effectiveness of \texttt{AgentAsk} in improving the performance of multi-agent systems under unchanged orchestration conditions.

\subsection{RQ2: Efficiency under Practical Budgets}
\label{app:rq2}
Tables~\ref{tab:gptswarm_all}, \ref{tab:maas_all}, and \ref{tab:masrouter_all} summarize the performance of \texttt{AgentAsk}, \texttt{+GPT-5}, and \texttt{origin} across five benchmarks. The results demonstrate that \texttt{AgentAsk} not only outperforms the original setting but also remains much more efficient compared to \texttt{+GPT-5}. Specifically, \texttt{AgentAsk} reduces the computational overhead, as evidenced by its lower extra costs, while maintaining similar or better accuracy across all frameworks. For example, in the GPTSwarm framework, \texttt{AgentAsk} achieves a +2.32\%~+4.34\% improvement in accuracy compared to the baseline while reducing extra costs significantly. This confirms that \texttt{AgentAsk} provides a practical and efficient solution that achieves competitive accuracy while staying within practical operational budgets.

\begin{figure}[ht]
    \centering
    \includegraphics[width=\linewidth]{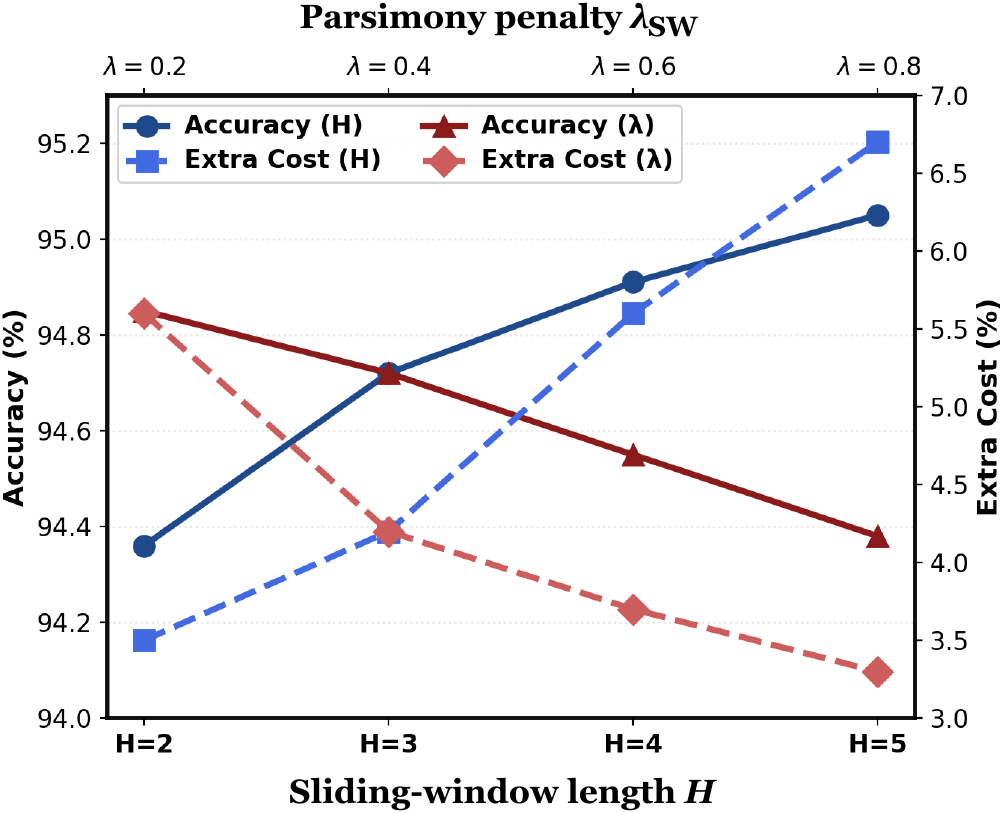}
    \caption{Sensitivity to window $H$ and penalty $\lambda_{\mathrm{sw}}$, highlighting a stable region near the default and the accuracy–efficiency trade-off.}
    \label{fig:sensitivity}
\end{figure}

\subsection{RQ3: Ablations and Robustness}
\label{app:rq3}
Table~\ref{tab:ablation_latency_extra} presents the results of an ablation study that investigates the impact of different reward components (\(r^{par}\), \(r^{eff}\), \(r^{fmt}\)) on the performance of \texttt{AgentAsk}. The ablation results highlight the importance of each reward component in balancing performance (accuracy) and efficiency (latency and extra cost). Removing any of the reward components leads to a noticeable drop in accuracy and an increase in latency or extra cost. For instance, excluding \(r^{par}\) results in a decrease of 0.20\% in accuracy and an increase in latency by 8\%, demonstrating the necessity of including all reward components for optimal performance.

Additionally, we compare the performance of \texttt{AgentAsk} with other reinforcement learning-based methods such as PPO and GRPO across five benchmarks, as shown in Table~\ref{tab:ppo_grpo_ours}. The results clearly indicate that \texttt{AgentAsk} consistently outperforms both PPO and GRPO in terms of accuracy, while maintaining a relatively low latency and extra cost. In particular, on the MMLU and GSM8K benchmarks, \texttt{AgentAsk} achieves significant improvements in accuracy (+2.25\% and +3.18\%, respectively), highlighting its robustness and effectiveness. These findings demonstrate that \texttt{AgentAsk} provides a reliable and efficient solution across different configurations and benchmarks, offering a promising approach for error mitigation in multi-agent systems. Finally, Figure~\ref{fig:sensitivity} illustrates the sensitivity of \textbf{AgentAsk} to the window size \(H\) and the parsimony weight \(\lambda_{\mathrm{sw}}\), showing that the clarifier is robust to parameter variations, with stable performance across a range of settings. This confirms that \textbf{AgentAsk} is both effective and adaptable.

\begin{figure*}[!t]
    \centering
    \includegraphics[width=0.85\linewidth]{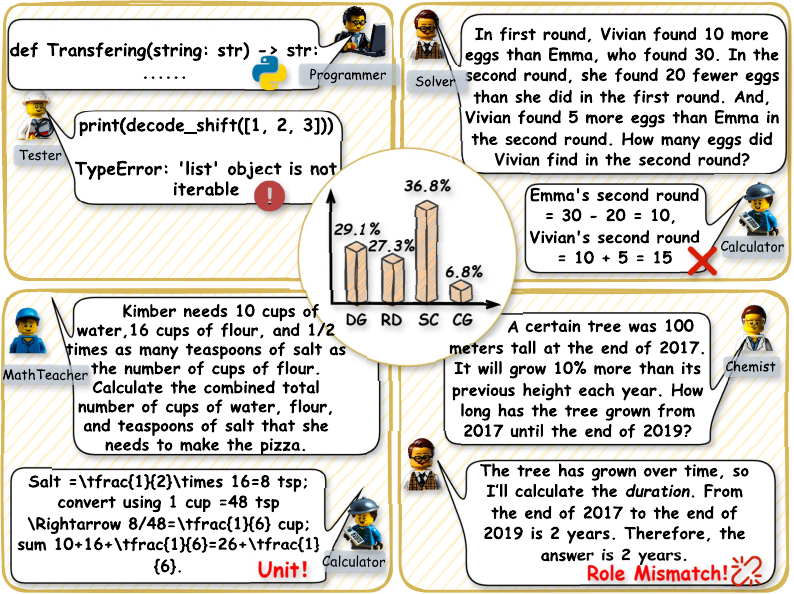}
    \caption{\textbf{The case of our error taxonomy.} In the middle shows the fraction of the four types of errors.}
    \label{apppic:taxonmycaseepic}
\end{figure*}

\section{Case Study of Error Taxonomy}
\label{app:case}

The following traces provide concrete grounding for the edge-level error taxonomy established in Section~\ref{taxonomy}. Each case is presented in a consistent four-column format as shown in Tables~\ref{tab:appendix_agentask_dg_single_case}, \ref{tab:appendix_agentask_sc_two_rows_inner_vlines_query}, \ref{tab:appendix_agentask_rd_single_case}, and \ref{tab:appendix_agentask_cg_two_cases}, detailing the \textbf{Query}, the original agent output (\textbf{Origin}), the clarification intervention by \textbf{AgentAsk}, and the resulting \textbf{Revised Message}. Figure~\ref{apppic:taxonmycaseepic} provides a visual overview and index of our edge-level error taxonomy. Its four quadrants each sketch a minimal, real trace exemplifying one of the four core error types (\textit{i.e.}, DG, RD, SC, and CG), while the central donut reports their prevalence in our annotated logs ($N$=824). This figure serves as a conceptual guide and navigational aid, directing the reader to the corresponding comprehensive case studies in Tables~\ref{tab:appendix_agentask_dg_single_case}--\ref{tab:appendix_agentask_cg_two_cases}. These tables provide the complete, granular detail of each interaction, including the original faulty message, the precise clarifying question posed by \textit{AgentAsk}, and the subsequent corrected handoff, enabling a thorough examination of the error mitigation process. 
\subsection{Data Gap (DG)}

Missing details at an edge force downstream agents to guess, which opens a cascade. The clarification asks for the exact boundary behavior that is absent at \(x_t\) and removes failure before the message flows further. The details can be seen in Table~\ref{tab:appendix_agentask_dg_single_case}.

\begin{table*}[!t]
\centering
\renewcommand{\arraystretch}{1.12}
\setlength{\tabcolsep}{6pt}

\caption{\textbf{AgentAsk Case (Data Gap).} The table shows one code-oriented edge where an upstream message omits boundary behavior: the query asks for a shortest-palindrome constructor but the origin message assumes a non-empty string and indexes without guards, creating an empty-input gap. Columns list the original query, the erroneous origin message, the minimal edge-level Ask (what/whom to ask), and the corrected handoff that adds explicit length checks and a safe return for the empty string.}
\label{tab:appendix_agentask_dg_single_case}
\begin{tabular}{p{0.22\textwidth}|p{0.22\textwidth}|p{0.22\textwidth}|p{0.22\textwidth}}
\hline
\rowcolor{CadetBlue!16}
\textbf{Query} & \textbf{Origin (Data Gap)} & \textbf{AgentAsk} &\textbf{Revised Message} \\
\hline

\begin{minipage}[t]{\linewidth}\vspace{2pt}
\begingroup
\lstset{
  language=Python,
  basicstyle=\ttfamily\scriptsize,
  columns=fullflexible,
  keepspaces=true,
  showstringspaces=false,
  breaklines=true,
  breakatwhitespace=true,
  frame=single
}
\begin{lstlisting}[linewidth=\linewidth]
def is_palindrome(string: str) -> bool:
    """Test if given string is a palindrome."""
    return string == string[::-1]

def make_palindrome(string: str) -> str:
    """Find the shortest palindrome that begins with a supplied string.
    Algorithm idea:
    - Find the longest postfix of supplied string that is a palindrome.
    - Append to the end the reverse of the prefix before that postfix.
    >>> make_palindrome('')
    """
\end{lstlisting}
\endgroup
\end{minipage}
&
\begin{minipage}[t]{\linewidth}\vspace{0pt}\itshape
I will implement make\_palindrome directly from the description. If the first and last characters match and the whole string is already a palindrome, I can immediately return it. Otherwise, I scan from right to left to detect the longest palindromic suffix and then append the reverse of the missing prefix. Concretely, I start with if string[0] == string[-1] and is\_palindrome(string): return string, then proceed into a loop to locate the suffix boundary. This covers typical inputs like words and short phrases. The docstring examples highlight the construction for ordinary strings, so I will assume the input is a standard non-empty string and does not require a guard for corner inputs such as the empty string or single-character strings. With that in mind, I will finalize the suffix search and produce the shortest palindrome output without additional checks or preconditions.
\\
\end{minipage}
&
\begin{minipage}[t]{\linewidth}\vspace{0pt}
\textbf{AgentAsk} $\to$ \textbf{ProgrammingExpert}:\\[-1pt]
\textit{“Should the function define behavior for the empty string explicitly  and add a quick guard for length 0 or 1 before indexing \texttt{string[0]}?”}\\[4pt]
\includegraphics[width=\linewidth]{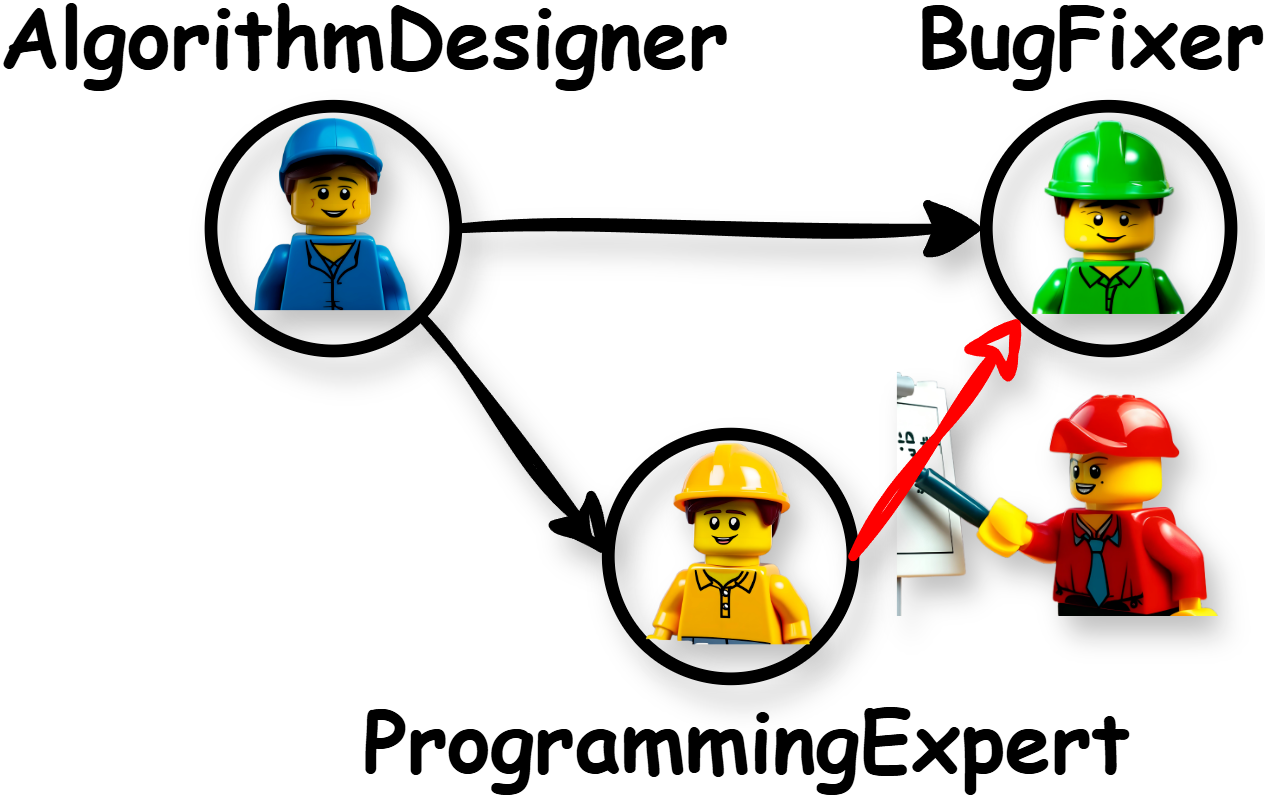}
\end{minipage}
&
\begin{minipage}[t]{\linewidth}\vspace{0pt}\itshape
The current logic indexes string[0] and string[-1] without guarding for length, which will raise IndexError on the empty string and is unnecessary on length-1 inputs. I will add explicit handling up front: if len(string) == 0, return ''; if len(string) == 1 or is\_palindrome(string), return string. Only for longer inputs will I search for the longest palindromic suffix and append the reverse of the missing prefix. This preserves correctness on normal cases while ensuring the edge case is well-defined and safe.
\end{minipage}
\\
\hline
\end{tabular}
\end{table*}

\subsection{Signal Corruption (SC)}

A wrong conversion or reference enters the chain and is relayed unchanged. The clarification pins down the intended unit or base quantity, which stops the propagation locally and earns positive effectiveness per Eq.~\eqref{eq:rl_eff}. Results are summarized in Table~\ref{tab:appendix_agentask_sc_two_rows_inner_vlines_query}.

\begin{table*}[!t]
\centering
\renewcommand{\arraystretch}{1.12}
\setlength{\tabcolsep}{6pt}

\caption{\textbf{AgentAsk Case (Signal Corruption)}.Two independent cases illustrate how a wrong conversion or misapplied base quantity can propagate along an edge. Case 1 (pizza units): the origin message forces unit unification and converts teaspoons to cups unasked, whereas the Ask clarifies “report as listed,” yielding 26 cups + 8 teaspoons (34 as count of listed quantities). Case 2 (reading remainder): the origin treats Tuesday’s \(\tfrac{1}{4}\) as a fraction of the whole; the Ask pins it to the Monday remainder and yields 60 pages left.  }
\label{tab:appendix_agentask_sc_two_rows_inner_vlines_query}
\begin{tabular}{p{0.19\textwidth}|p{0.25\textwidth}|p{0.22\textwidth}|p{0.23\textwidth}}
\hline
\rowcolor{CadetBlue!16}
\textbf{Query} & \textbf{Origin (Signal Corruption)} & \textbf{AgentAsk} & \textbf{Revised Message} \\
\hline

\begin{minipage}[t]{\linewidth}\vspace{0pt}\itshape
To make pizza, together with other ingredients, Kimber needs 10 cups of water, 16 cups of flour, and 1/2 times as many teaspoons of salt as the number of cups of flour. Calculate the combined total number of cups of water, flour, and teaspoons of salt that she needs to make the pizza.
\end{minipage}
&
\begin{minipage}[t]{\linewidth}\vspace{0pt}\itshape
We have 10 cups of water and 16 cups of flour. Salt equals half as many teaspoons as flour cups, so that is \( \tfrac{1}{2}\times 16 = 8 \) teaspoons. Because the prompt asks for a combined total in cups, I will convert teaspoons to cups and sum in a single unit for uniformity. Using \(1\,\mathrm{cup}=48\,\mathrm{tsp}\), we get \(8/48=1/6\) cup \(\approx 0.1667\) cup. Hence, the total is \(10+16+\tfrac{1}{6}=26\tfrac{1}{6}\) cups \(\approx 26.1667\) cups. I will report the answer in cups only to avoid mixing units.
\\
\end{minipage}
&
\begin{minipage}[t]{\linewidth}\vspace{0pt}
\textbf{AgentAsk $\to$ AlgorithmEngineer:}\\[-1pt]
\textit{“Should teaspoons be converted to cups before calculating the total?”}\\[4pt]
\includegraphics[width=\linewidth]{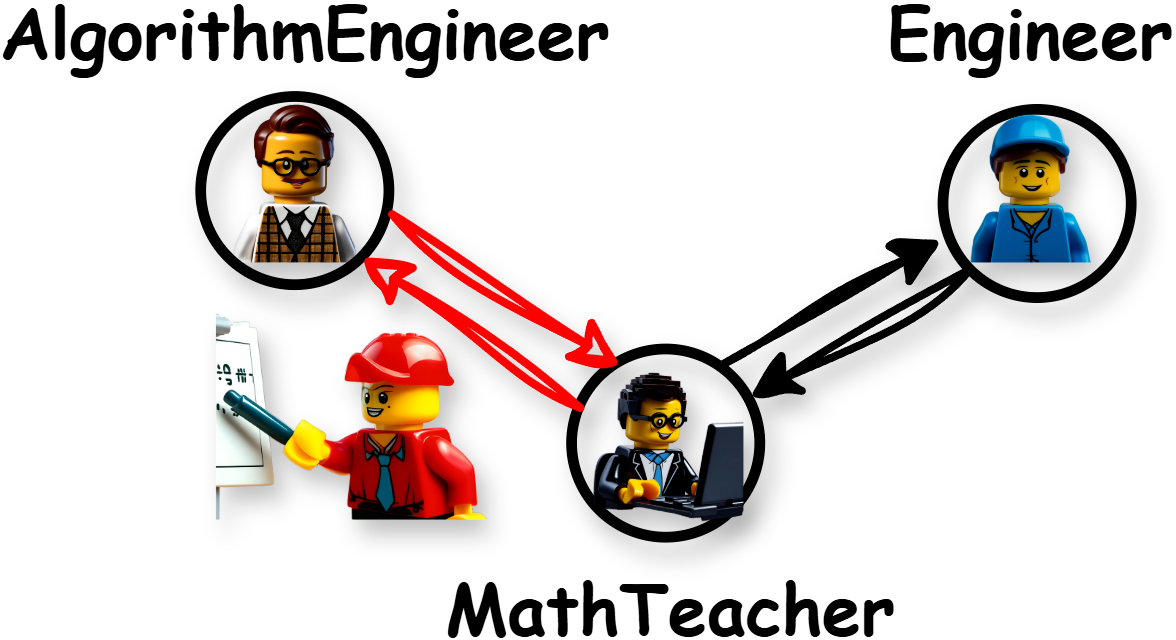}
\end{minipage}
&
\begin{minipage}[t]{\linewidth}\vspace{0pt}\itshape
 The prompt specifies cups for water and flour and teaspoons for salt without requiring unification. To avoid injecting an unstated policy, I will preserve units as given and compute the combined total \emph{as listed}. Quantities: water \(=10\) cups, flour \(=16\) cups, salt \(=8\) teaspoons. Reporting the combined count of listed quantities gives 34 \((26+8)\). %If a single-unit answer is later requested, we can convert 8 teaspoons to \(1/6\) cup and additionally report \(26\tfrac{1}{6}\) cups.
\end{minipage}
\\
\hline

\begin{minipage}[t]{\linewidth}\vspace{0pt}\itshape
A book has 120 pages. On Monday, a student reads one-third of the book. On Tuesday, the student reads one quarter of the \emph{remainder}. How many pages are left unread after Tuesday?
\end{minipage}
&
\begin{minipage}[t]{\linewidth}\vspace{0pt}\itshape
 I will compute the total pages read by treating both fractions as portions of the whole book for simplicity: Monday \(= \tfrac{1}{3}\times 120 = 40\) pages; Tuesday \(= \tfrac{1}{4}\times 120 = 30\) pages. Total read \(= 70\) pages, so pages left \(= 120-70=50\). I will pass along “50 pages left” as the final quantity.
\end{minipage}
&
\begin{minipage}[t]{\linewidth}\vspace{0pt}
\textbf{AgentAsk $\to$ MathSolver:}\\[-1pt]
\textit{“Does ‘one quarter’ on Tuesday refer to the remainder after Monday, rather than the whole book?”}\\[4pt]
\includegraphics[width=\linewidth]{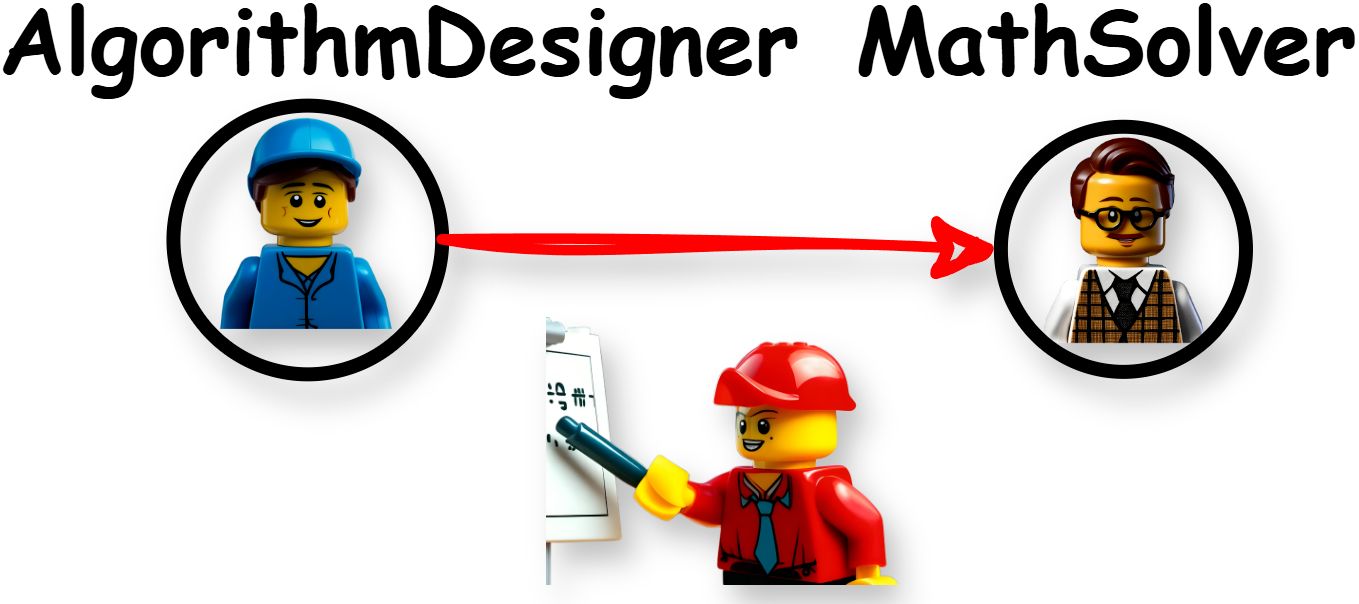}
\end{minipage}
&
\begin{minipage}[t]{\linewidth}\vspace{0pt}\itshape
Thanks for the clarification. “One quarter” is with respect to the remainder after Monday. Compute precisely: Monday reads \( \tfrac{1}{3}\times 120 = 40\) pages; remainder \(= 120-40 = 80\) pages. Tuesday reads \( \tfrac{1}{4}\times 80 = 20\) pages. Total read \(= 40+20=60\); therefore pages left \(= 120-60 = 60\). I will correct the intermediate assumption and pass along “60 pages left” as the final quantity.\\
\end{minipage}
\\
\hline
\end{tabular}
\end{table*}

\subsection{Referential Drift (RD)}

Ambiguous symbols drift across turns and break consistency. The clarification fixes names and indices at the edge and aligns all relations, after which the downstream solution becomes stable. The details can be seen in Table~\ref{tab:appendix_agentask_rd_single_case}.

\begin{table*}[!t]
\centering
\renewcommand{\arraystretch}{1.12}
\setlength{\tabcolsep}{6pt}

\caption{\textbf{AgentAsk Case (Referential Drift).}A single math word-problem case where ambiguous references (\(T, E\) with dropped round indices) cause symbol drift across turns and wrong relations like \(T_2=E_1-20\). The Ask enforces explicit indexing (\(T_1,T_2,E_1,E_2\)) and confirms the intended relations (\(T_2=T_1-20\), \(E_2=2T_2\)), after which the downstream solution stabilizes . }
\label{tab:appendix_agentask_rd_single_case}
\begin{tabular}{p{0.20\textwidth}|p{0.24\textwidth}|p{0.22\textwidth}|p{0.24\textwidth}}
\hline
\rowcolor{CadetBlue!16}
\textbf{Query} & \textbf{Origin (Referential Drift)} & \textbf{AgentAsk} &\textbf{ Revised Message} \\
\hline

\begin{minipage}[t]{\linewidth}\vspace{0pt}
\textit{In April, Tank gathered 10 more Easter eggs than Emma in their first round of egg hunt. Emma gathered twice as many eggs as Tank in their second round, while Tank's second-round total was 20 less than in the first round. The combined pile with 6 other people was 400 eggs, and Emma gathered 60 eggs in the second round. Find how many eggs the 6 other egg hunters collected.}
\end{minipage}
&
\begin{minipage}[t]{\linewidth}\vspace{0pt}

\textit{\textbf{MathAnalyst} $\rightarrow$ \textbf{Mathematician}: Let \(T\) and \(E\) denote Tank and Emma. I will use \(T_1,E_1\) for round one and \(T_2,E_2\) for round two, but for brevity I may drop the subscripts. From the text, “Tank's second round was 20 less than she had gathered in the first round,” so the difference between Tank’s second and first rounds is 20. Emma’s second round is “twice as many as Tank in their second round,” which suggests a doubling relative to Tank's performance. Since Emma has 60 in the second round, we may set \(E_2=60\). Because \(E_1 - T_2 = 20\) , we can write \(T_2 = E_1 - 20\). Also, “Tank gathered 10 more than Emma in their first round” gives \(T_1 = E_1 + 10\). With these, the total across two rounds plus the six others should reach 400; I pass these relations forward for solution.}
\\
\end{minipage}
&
\begin{minipage}[t]{\linewidth}\vspace{0pt}
\textbf{AgentAsk} $\to$ \textbf{MathAnalyst}:\\[-1pt]
\textit{“Please confirm unambiguous referents: Does ‘20 less’ apply to Tank’s own rounds \((T_2 = T_1 - 20)\)? Does ‘twice as many’ mean \(E_2 = 2\,T_2\)? }\\[4pt]
\includegraphics[width=\linewidth]{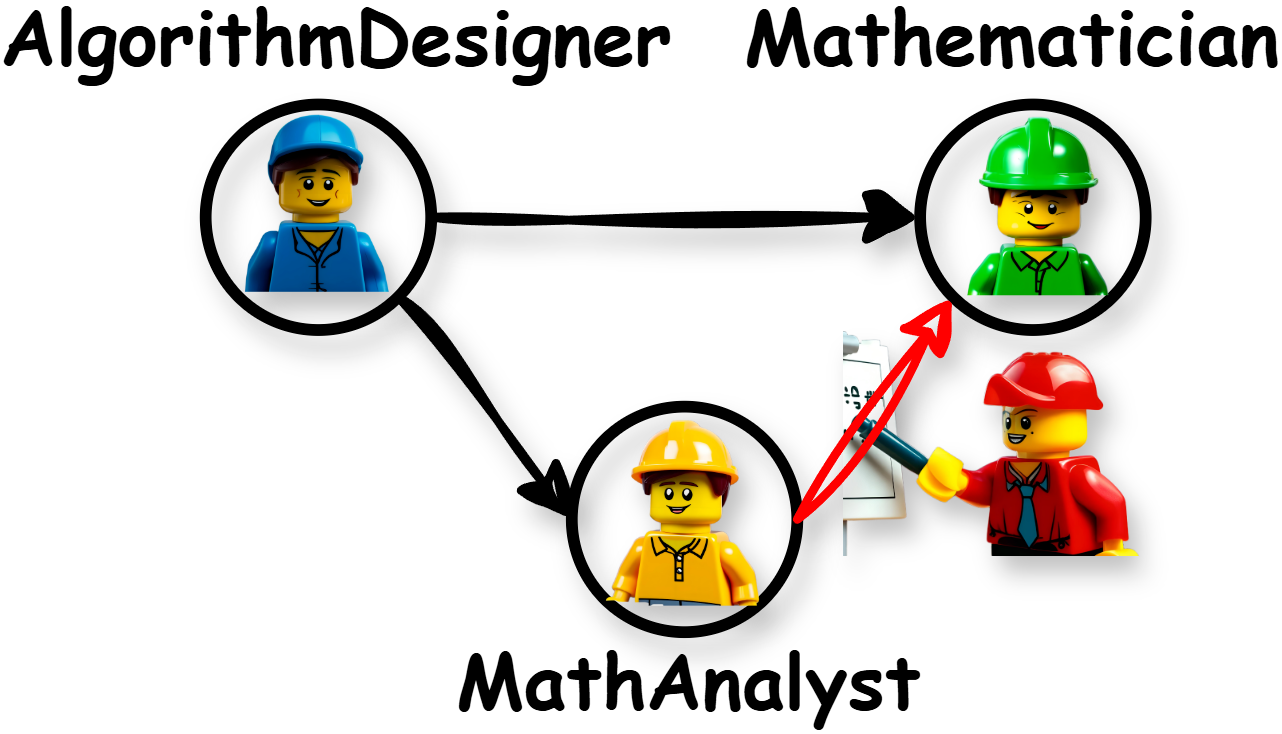}
\end{minipage}
&
\begin{minipage}[t]{\linewidth}\vspace{0pt}\itshape
 Fix symbols explicitly: \(T_1,T_2\) for Tank (rounds 1,2) and \(E_1,E_2\) for Emma (rounds 1,2). Map statements to equations: \(T_1 = E_1 + 10\); \(T_2 = T_1 - 20\); \(E_2 = 2\,T_2\); and the given \(E_2 = 60\). From \(E_2 = 60\) we get \(T_2 = 30\). Then \(T_1 = 50\). Using \(T_1 = E_1 + 10\) gives \(E_1 = 40\). The subtotal for Tank and Emma is \(T_1 + T_2 + E_1 + E_2 = 50 + 30 + 40 + 60 = 180\). Therefore, the six other hunters collected \(400 - 180 = 220\) eggs. The earlier drift came from writing \(T_2 = E_1 - 20\), which incorrectly bound the “20 less” clause to Emma’s first round; keeping indices attached eliminates the ambiguity and yields a consistent solution.
\\
\end{minipage}
\\
\hline
\end{tabular}
\end{table*}

\subsection{Capability Gap (CG)}
The role at the edge lacks the skill required by the subtask. The clarification requests a reroute to a capable recipient and specifies the expected output form, which resolves the mismatch with a single step. Table~\ref{tab:appendix_agentask_cg_two_cases} shows the corrected messages.

\begin{table*}[t]
\centering
\renewcommand{\arraystretch}{1.12}
\setlength{\tabcolsep}{6pt}

\caption{\textbf{AgentAsk Case (Capability Gap).}Two cases where the role at the edge is misaligned with the subtask. Case 1 (tree growth): an Economist returns elapsed time (“2 years”) instead of the requested growth amount; the Ask reroutes to AlgorithmEngineer, who computes compound growth (21 meters). Case 2 (coding): a non-coding role provides a qualitative description instead of an implementation; the Ask redirects to ProgrammingExpert, who supplies a correct function and complexity note.}
\label{tab:appendix_agentask_cg_two_cases}
\begin{tabular}{p{0.16\textwidth}|p{0.26\textwidth}|p{0.22\textwidth}|p{0.26\textwidth}}
\hline
\rowcolor{CadetBlue!16}
\textbf{Query} & \textbf{Origin (Capability Gap)} & \textbf{AgentAsk} & \textbf{Revised Message} \\
\hline

\begin{minipage}[t]{\linewidth}\vspace{0pt}
\textit{A certain tree was 100 meters tall at the end of 2017. It will grow 10\% more than its previous height each year. How long has the tree grown from 2017 until the end of 2019?} \\
% Reasoning: FullConnected \\
% Roles: [AlgorithmEngineer, Scientist, Economist, Engineer]
\end{minipage}
&
\begin{minipage}[t]{\linewidth}\vspace{0pt}\itshape
\textbf{Economist} $\rightarrow$ \textbf{Engineer}: The question asks about the period from 2017 to the end of 2019, which spans two calendar years. Interpreting “how long has the tree grown” as the elapsed time, we simply count the years: 2018 and 2019. Therefore, the tree has grown for a total of 2 years. The relevant arithmetic is duration-based rather than quantitative growth, so the final answer is 2, reflecting the time interval rather than the magnitude of height change. 
\end{minipage}
&
\begin{minipage}[t]{\linewidth}\vspace{0pt}
\textbf{AgentAsk} $\to$ \textbf{AlgorithmEngineer}:\\[-1pt]
\textit{“The user asks for the \emph{growth amount} (height increase), not the \emph{elapsed time}. Please route to AlgorithmEngineer or Scientist for compound-growth computation.”}\\[4pt]
\includegraphics[width=\linewidth]{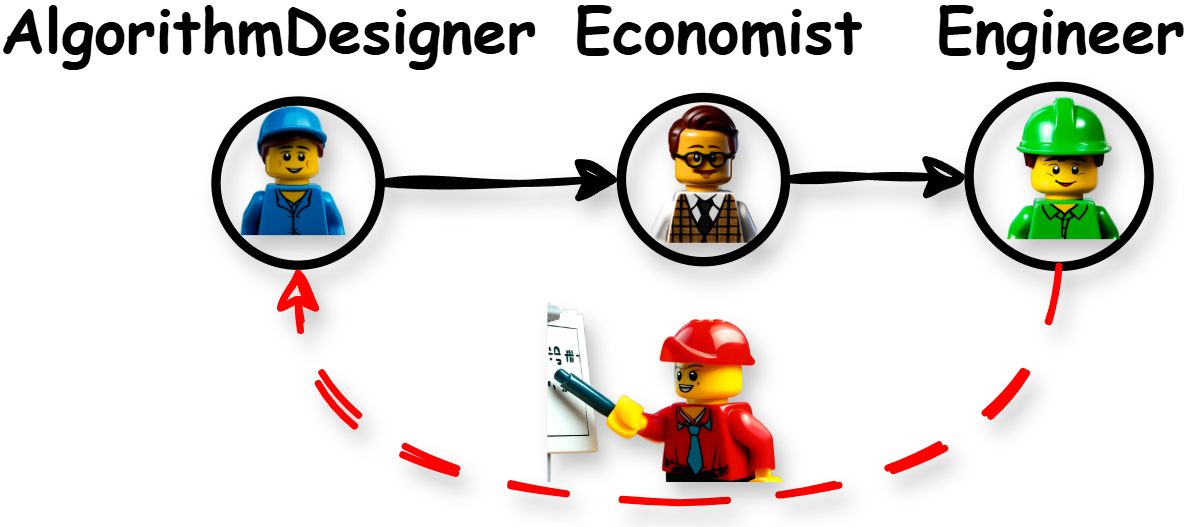}
\end{minipage}
&
\begin{minipage}[t]{\linewidth}\vspace{0pt}\itshape
\textbf{AlgorithmEngineer}: We compute the compound growth from end-2017 baseline \(H_0=100\). Each year multiplied by \(1.10\). After two years: \(H_2=100\times(1.10)^2=121\) meters. The growth amount is \(\Delta H = H_2-H_0 = 121-100=21\) meters. Thus, the correct quantity is a height difference, not a duration; the previously provided “2 years” reflects a capability mismatch rather than the requested quantitative outcome. Final: the tree has grown by 21 meters over the period.
\end{minipage}
\\
\hline

\begin{minipage}[t]{\linewidth}\vspace{0pt}
\textit{Write a function sum\_even\_squares (n: int) -> int that returns the sum of squares of all even integers from 1 to \texttt{n} inclusive. Provide a concise implementation and explain its complexity.}\\

\end{minipage}
&
\begin{minipage}[t]{\linewidth}\vspace{0pt}\itshape
\textbf{MathAnalyst} $\rightarrow$ \textbf{Scientist}: To address the request, we consider how even numbers behave in general and recall that squaring preserves parity. The sum can conceptually be obtained by repeatedly adding the square of each even number up to the bound. The main idea is to iterate across the relevant subset and aggregate the contributions. From a scientific perspective, the computation is straightforward and should be feasible, though nuances about performance might emerge for very large inputs. In summary, the result equals the accumulated squared values of all even integers no greater than \texttt{n}.
\end{minipage}
&
\begin{minipage}[t]{\linewidth}\vspace{0pt}
\textbf{AgentAsk} $\to$ \textbf{Engineer}:\\[-1pt]
\textit{“The user explicitly requests an implementation and complexity discussion.  Ensure the function handles small \texttt{n}.”}\\[4pt]
\includegraphics[width=\linewidth]{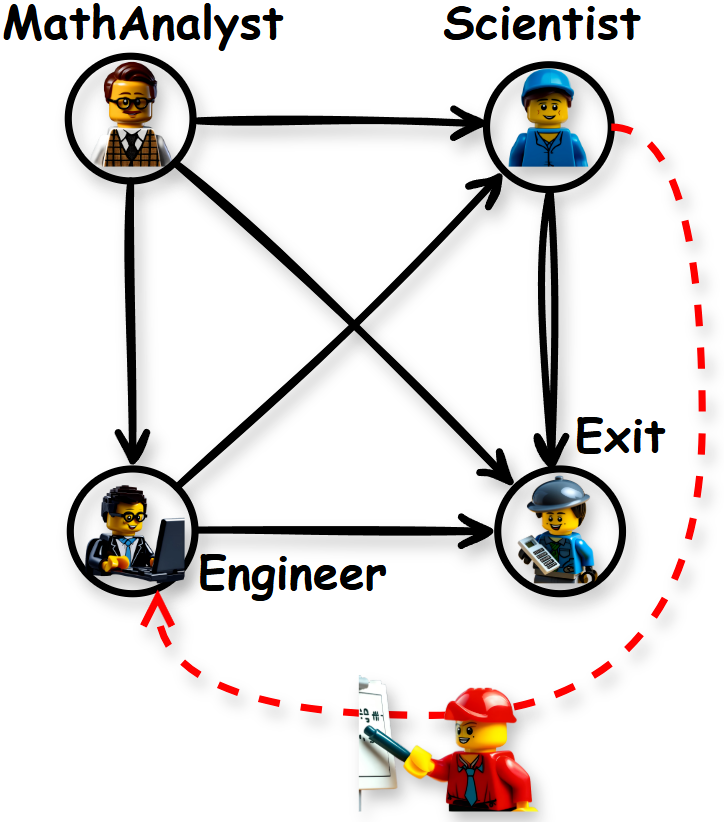}
\end{minipage}
&
\begin{minipage}[t]{\linewidth}\vspace{0pt}
\begingroup
\lstset{
  language=Python,
  basicstyle=\ttfamily\scriptsize,
  columns=fullflexible,
  keepspaces=true,
  showstringspaces=false,
  breaklines=true,
  breakatwhitespace=true,
  frame=single
}
\begin{lstlisting}[linewidth=\linewidth]
def sum_even_squares(n: int) -> int:
    """Return sum of k^2 for even k in [1..n]."""
    if n < 2:
        return 0
    # Iterate by step 2 over even numbers only
    return sum(k*k for k in range(2, n+1, 2))
\end{lstlisting}
\endgroup
\itshape
ProgrammingExpert: The function iterates over even integers only, avoiding unnecessary checks, and sums their squares. For example, \texttt{n=6} uses \(\{2,4,6\}\) giving \(4+16+36=56\). Time complexity is \(O(n/2)=O(n)\) with \(O(1)\) auxiliary space. A closed-form alternative via \(\sum (2i)^2 = 4\sum i^2\) for \(i=1.. \lfloor n/2 \rfloor\) can reduce iteration count with arithmetic. This reroute resolves the capability gap by assigning code synthesis to the appropriate role.\\
\end{minipage}
\\
\hline
\end{tabular}
\end{table*}

\section{Notations}
\label{sec:notations}
The symbols used in the text are explained in detail in the Table~\ref{tab:notations}.

\begin{table*}[!t]
\centering
\small
\renewcommand{\arraystretch}{1.2}
\setlength{\tabcolsep}{8pt}
\resizebox{\textwidth}{!}{
\begin{tabular}{@{}ll@{}}
\toprule
\textbf{Symbol} & \textbf{Description} \\ \midrule
$N$ & The total number of examples in the training dataset. \\
$x_i$ & The feature vector representing the state at edge $i$. \\
$y_i$ & The intervention decision at edge $i$, encoding whether to ask, whom to ask, and what to ask. \\
$\mathcal{D}$ & The training dataset consisting of edge-local examples $(x_i, y_i)$. \\
$\mathcal{L}_{\text{Total}}$ & The total training loss, combining $\mathcal{L}_{\text{type}}$ and $\mathcal{L}_{\text{ask}}$ . \\
$\lambda_{\text{ask}}$ & A parameter balancing the two components of the loss function. \\
$x_t$ & The edge-local state at time step $t$ during the message handoff. \\
$a_t$ & The action at time step $t$, consisting of a binary gate $z_t$, an agent selection $\tilde{v}_t$, and the clarification question $q_t$. \\
$z_t$ & A binary gate indicating whether to ask clarification at time step $t$. \\
$\tilde{v}_t$ & The agent selected to receive the clarification at time step $t$. \\
$q_t$ & The generated clarification question at time step $t$. \\
$\pi_\theta$ & The policy that governs the action sequence across an input, parameterized by $\theta$. \\
$\tau$ & A trajectory, which is a sequence of edge-local states and actions $\{(x_t, a_t)\}_{t=1}^T$. \\
$U(\tau)$ & The utility function measuring task success. \\
$C(\tau)$ & The cost incurred from clarifications during the task. \\
$B$ & The predefined budget limit for clarifications. \\
$r_t^{\mathrm{edge}}$ & The aggregated edge reward. \\
$R$ & The system-level reward, representing the final task success. \\
$A_t^{\mathrm{loc}}$ & The local advantage signal at edge $t$, calculated from the edge reward $r_t^{\mathrm{edge}}$. \\
$A_t^{\mathrm{glob}}$ & The global advantage signal at edge $t$, incorporating global feedback. \\
$w_t$ & The weight associated with the global advantage signal at time step $t$. \\
$b$ & The baseline for the global advantage signal. \\
$\lambda_R(t)$ & The time-varying weight controlling the integration of global feedback in the learning process. \\
$\rho_t$ & The ratio of probabilities used in the policy update during reinforcement learning. \\
$\beta$ & The penalty parameter for regularizing the policy updates. \\
$G = (V, E)$ & A message graph, where $V$ represents the set of nodes (agents) and $E$ represents the set of edges (message handoffs). \\
$\{e_t\}_{t=1}^{T}$ & A sequence of message handoffs, where each $e_t = (u_t \rightarrow v_t)$ denotes a message passing from agent $u_t$ to agent $v_t$. \\
$x^{\mathrm{in}}_t$ & The user query or subgoal at time step $t$. \\
$m_t$ & The message from agent $u_t$ at time step $t$. \\
$h_t$ & The encoded recent local interaction context at time step $t$. \\
$p^{\text{type}}_\theta(t_i \mid x_i)$ & The probability of the ask decision at edge $i$, given the context $x_i$. \\
$p^{\text{addr}}_\theta(v_i \mid x_i, t_i)$ & The probability of selecting the addressee $v_i$ for clarification at edge $i$. \\
$p^{\text{txt}}_\theta(q_i \mid x_i, t_i)$ & The probability distribution over clarification text $q_i$ at edge $i$. \\
$m_i$ & An indicator for whether a clarification is present at edge $i$. \\ \bottomrule
\end{tabular}
}
\caption{\textbf{Explanation of the notation and symbols used in this paper.}}
\label{tab:notations}
\end{table*}

\section{Prompt}
\label{sec:Prompt}

The complete instruction used by the clarifier is delineated in Figure~\ref{fig:agentask_prompt_box}. 
It explicitly specifies \textit{when to ask}, \textit{what to ask}, and \textit{whom to ask}, while keeping questions short and cost-aware.Due to the inconsistency in the framework, there may be slight differences in the input section of the Prompt, but this does not affect the overall consistency.

\begin{figure*}[!t]
\centering
\begin{minipage}{0.97\textwidth}
\begin{mdframed}[linewidth=0.6pt,roundcorner=6pt,backgroundcolor=gray!3]
\small
You are AgentAsk, an edge-level clarifier between two agents. Your job is to decide whether a minimal clarification should be asked before the message is handed off, so that small mistakes do not spread. If you do ask, decide what to ask, whom to ask (which agent in MAS), and write one short, concrete question. Keep it brief and low-cost. Do not change the overall workflow.

\medskip
Use the following error taxonomy when asking. Choose one type if you decide to ask; if no clarification is needed, set type="NONE" and leave the question empty.

\medskip
\textbf{DG — Data Gap} \\
Some required detail is missing, and the next agent would have to guess. \\
Signals: missing boundary cases; absent IDs, keys, or columns; unclear ranges or formats; placeholders. \\
How to ask: request the smallest missing piece in one short question. \\
Who to ask: usually the sender. \\
Do not ask: when the value is obvious from context or does not affect the outcome.

\medskip
\textbf{SC — Signal Corruption} \\
An intermediate value, unit, scale, or structure is wrong or malformed and may be copied forward as truth. \\
Signals: unit mismatch; off-by-one indices; broken JSON or tables; inconsistent time zones; impossible magnitudes. \\
How to ask: point to the exact field or structure and confirm or repair it. \\
Who to ask: usually the sender; the other agent in MAS if they must choose a canonical unit. \\
Do not ask: when downstream already normalizes it deterministically.

\medskip
\textbf{RD — Referential Drift} \\
Names or symbols do not refer to the same thing across turns, so agents bind to different entities. \\
Signals: pronouns like it or they; reused symbols without scope; conflicting aliases; index shifts; unlabeled columns. \\
How to ask: fix a single binding with a clear choice. \\
Who to ask: usually the sender; the other agent in MAS if they must commit to a binding for later steps. \\
Do not ask: when the binding is unambiguous from nearby context.

\medskip
\textbf{CG — Capability Gap} \\
The current addressee lacks the skill or role to complete the step. \\
Signals: narrative text where computation is needed; missing tool access; math or code assigned to a planner; required API not available. \\
How to ask: propose a minimal reroute or ask for the needed computation in one line. \\
Who to ask: the other agent in MAS being rerouted. \\
Do not ask: when a tiny hint lets the current role finish; in that case consider DG, RD, or SC first.

\medskip
\textbf{NONE — no ask} \\
Choose NONE when a single short question will not meaningfully reduce uncertainty, when the handoff is already sufficient, when policy or privacy would block the question, or when the next agent can fix it deterministically without new information. When using NONE, set to\_agent to null and leave question empty. Output only the required JSON.

\medskip
Here is the input of upstream agent ......
\end{mdframed}
\end{minipage}
\caption{\textbf{Edge-level clarifier prompt.} The box defines the instruction used by the clarifier to decide whether to ask, what to ask, whom to ask, and how to phrase a single, minimal question under budget constraints.}
\label{fig:agentask_prompt_box}
\end{figure*}

\end{document}